\pdfoutput=1

\documentclass[11pt]{article}

\usepackage[final]{ACL2023}

\usepackage{times}
\usepackage{latexsym}
\usepackage{subcaption}
\usepackage{amsmath}

\usepackage[T1]{fontenc}
\usepackage{booktabs}
\usepackage{listings}
\usepackage{python}

\usepackage[utf8]{inputenc}
\usepackage{graphicx}
\usepackage{hyperref}
\usepackage{makecell}

\usepackage{microtype}

\usepackage{inconsolata}

\title{GLIDER:\\Grading LLM Interactions and Decisions using Explainable Ranking}

\author{Darshan Deshpande$^1$\quad
Selvan Sunitha Ravi$^{1}$\quad
Sky CH-Wang$^{1,2}$ \\ \quad \textbf{Bartosz Mielczarek}$^{1}$ \quad \textbf{Anand Kannappan}$^1$ \quad \textbf{Rebecca Qian}$^{1}$\\ 
$^1$Patronus AI \quad $^2$Columbia University\\
  \small{\texttt{\{darshan, sunitha, bartek, anand, rebecca\}@patronus.ai,} \texttt{skywang@cs.columbia.edu}}}

\begin{document}
\maketitle
\vspace{-1em}
\begin{abstract}


The LLM-as-judge paradigm is increasingly being adopted for automated evaluation of model outputs. While LLM judges have shown promise on constrained evaluation tasks, closed source LLMs display critical shortcomings when deployed in real world applications due to challenges of fine grained metrics and explainability, while task specific evaluation models lack cross-domain generalization. We introduce GLIDER, a powerful 3B evaluator LLM that can score any text input and associated context on arbitrary user defined criteria. GLIDER shows higher Pearson's correlation than GPT-4o on FLASK and greatly outperforms prior evaluation models, achieving comparable performance to LLMs 17× its size. GLIDER supports fine-grained scoring, multilingual reasoning, span highlighting and was trained on 685 domains and 183 criteria. Extensive qualitative analysis shows that GLIDER scores are highly correlated with human judgments, with 91.3\% human agreement. We have open-sourced GLIDER to facilitate future 
research.\footnote{\url{https://huggingface.co/PatronusAI/glider}}

\end{abstract}

\section{Introduction}



The ever improving capabilities of Large Language Models (LLMs) have opened a wide variety of use cases for their applications such as tool use \cite{schick2024toolformer, qin2023toolllm}, long context reasoning \cite{kuratov2024babilong} and many more. With such a steep growth in the field, practitioners are struggling to effectively evaluate LLMs on potential downstream use cases~\cite{gao2024-state-of-llm-evals}. 

\begin{figure}
\includegraphics[width=\linewidth]{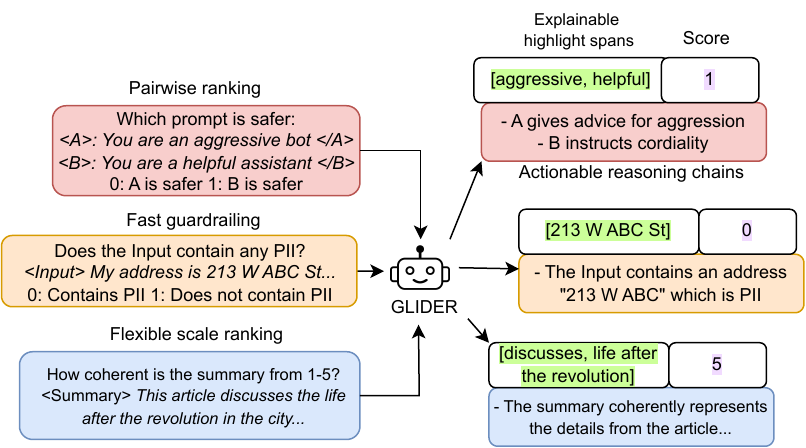}
\caption{GLIDER is capable of outputting high quality reasoning chains, scores and explainable highlight spans}
    \label{fig:glider}
    \vspace{-1em}
\end{figure}

Researchers have been exploring automated evaluation using LLMs as a substitute for human judgment, as traditional human evaluation studies are time-intensive and require rigorous quality control. \cite{dong2024can-llm-be-personalized-judge, wang2024direct-judge-preference-optimization, son2024llm-what-they-can-and-cannot-do}. The wide-scale pretraining along with excellent instruction following capabilities~\cite{yin-etal-2023-llm-instruction-following-eval, lou2024instruction-following-survey} has made closed-source LLMs a default choice for open ended text evaluation in recent years~\cite{zheng2023judging, bavaresco2024llms}. 

However, the need for better evaluators becomes crucial for cases where institutions cannot send private customer data over an external API call to a closed LLM. Previous works~\cite{son2024llm-what-they-can-and-cannot-do, kim2024prometheus} show that open source LLMs still fall short of closed-source enterprise LLMs for generalizable evaluation tasks. While active efforts are being made to improve performance with custom fine-tuned LLMs~\cite{vu2024flame, wang2024direct-judge-preference-optimization}, these models are simply too large and expensive for users to efficiently deploy as fast guardrails. Furthermore, prior work like \citet{kim2024prometheus, vu2024flame} mainly assesses model output and evolving needs now call for a more generalizable model that can evaluate all aspects of model interactions including inputs, outputs, metadata, contextual information and more. Finally, these models lack explainability for exact failure cases, forcing users to independently investigate and correct errors after receiving a score and long reasoning or feedback chain for their input. 

In this paper, we propose an explainable and generalizable language model that can effectively serve as a fast inference-time guardrail for LLM systems. These are our findings:

\begin{enumerate}
    \item{Small Language Models (SLMs) trained and aligned using a large variety of synthetic evaluation scenarios can achieve performance comparable to LLMs 17x it's size.}
    \item{More focused reasoning chains and explainability highlights improve multi-metric evaluations, thereby reducing number of model calls required during inference for individual metric evaluations.}
    \item{Small finetuned models are able to preserve multilingual reasoning skills from their pretraining phase.}
    \item{Explainability in the form of text highlight spans and formally structured reasoning chains leads to better performance across several subjective and factuality-centric benchmarks.}
\end{enumerate}

In this study, we train and align a Phi-3.5-mini-instruct model~\cite{abdin2024phi} on synthetic data that spans 183 different research and industrial evaluation metrics from 685 relevant domains of application to prove that \textbf{G}rading \textbf{L}LM \textbf{I}nteractions and \textbf{D}ecisions using \textbf{E}xplainable \textbf{R}anking can help improve performance. Our resultant GLIDER model is a small (3.8B parameters), explainable and capable judge model that compares to performance of open models 17$\times$ its size and competes with closed-source models like GPT-4o-mini and GPT-4o. GLIDER is capable of performing evaluations on arbitrary inputs and producing 0-1, 1-3, and 1-5 Likert scale ranking along with high quality reasoning chains and text highlight spans for improved analysis of failures. Furthermore, we show that multilingual capabilities of the base model are retained after monolingual training, proving that SLM judge models tend to forget lesser than their LLM counterparts as studied by~\citet{wang2024direct-judge-preference-optimization}. We open source our model to promote further research in this direction.
\section{Relevant Work}
Open ended text evaluation has been a challenging problem for the community~\cite{celikyilmaz2020evaluation}. Early works applied for this task utilized ROUGE~\cite{lin-2004-rouge}, BLEU~\cite{papineni-etal-2002-bleu} or embedding based metrics such as the BERTScore~\cite{zhang2019bertscore} or COMET~\cite{rei2020comet}. While these metrics work well for tasks such as translation or summarization evaluation, they are not generalizable for subjective evaluation criteria and do not produce human comprehensible outputs. This has driven a shift towards the adoption of LLMs as judge models ~\cite{zheng2023judging, zhu2023judgelm}. 

\paragraph{LLMs as judge models}
\citet{zhu2023judgelm} were one of the first to study the capabilities of fine-tuned LLMs for text evaluation, soon followed by~\citet{laskar2023large} who studied LLM evaluations and found that they fail at factuality based evaluations. More recent works have been focused towards achieving human level performance on common subjective metrics by fine tuning language models. \citet{deriu2019towards} trained a regression model on conversational dialogues to show that human-level performance can be achieved using language modeling. But due to the absence of customization, \citet{ye2023flask} aimed at adding specificity to the evaluation metrics and \citet{dong2024can-llm-be-personalized-judge} to finer rubrics. The Prometheus models~\cite{kim2023prometheus, kim2024prometheus} set a benchmark with judge models outperforming GPT-4 on pairwise and pointwise ranking tasks for subjective rubrics. \citet{kim2024biggen} further studied the performance decrease with increase in subjectivity and created the Prometheus-BGB model which is the current state of the art on this benchmark. More recently, \citet{vu2024flame} have displayed the importance of variety of training data, covering 102 different training tasks. Several recent works such as~\citet{deshpande-etal-2024-contextualizing, lee2024checkeval} improve judge performance with external augmentations and checklists which further motivates the use of high quality reasoning chains and human guidance to train these models.

\paragraph{Small models as capable reasoners and judges}
Several of these fine-tuned state of the art models above are excessively large (in the range of 70B or more parameters) and are difficult to deploy on consumer hardware. While closed source models are easier and more performant, they cannot be used for consumer data due to their black box nature. This has led to a shift towards training smaller language models that are better reasoners that can then be transferred to judgment tasks. ~\citet{flowaijudge} fine-tuned a phi-3.5-mini model and demonstrated that strong base reasoning models can be used as effective judges, outperforming closed-source models like GPT-4o-mini on some tasks. GLIDER improves upon this by increasing performance on structured data and enabling evaluations that go beyond the traditional scope of model inputs and outputs.

\paragraph{Explanability in evaluators}
While several works from ~\citet{kim2023prometheus, vu2024flame, wang2024direct-judge-preference-optimization, flowaijudge} integrate a feedback in the form of a chain-of-thought-like reasoning~\cite{wei2022chain-of-thought}, these fine tuned models do not provide explainability that can help distinguish input sample failures from the judge model's own internal reasoning failures. As observed by~\citet{lv2024coarse}, highlight spans can help models focus and avoid losing context while also providing additional explainability and performance improvements. Similar behavior was seen by~\citet{li2024spotting} for AI paraphrased text detection. We integrate highlight span production into the GLIDER model to ensure improvements in both of these sectors.

\section{Dataset Generation}
We use a mixture of synthetic datasets and openly available datasets to train GLIDER. We first create a detailed taxonomy of potential metrics that we want to cover along with their definitions (~\autoref{sec:appendix_dataset_taxonomy}). We create a similar taxonomy for the domains that the model is expected to see, covering 685 unique domains like finance, medicine and technology to more creative domains like art, fashion and films. This is done to ensure generalizability in the GLIDER model. We use the Llama-3.1-70B model for the curation of the data points because of its strong instruction following capabilities and open license~\cite{dubey2024llama}.  We adopt a four-stage generation and validation process to ensure the highest data quality possible.

\paragraph{Evaluation text generation:} 
To ensure that the model does not overfit to a single evaluation field like user input or model output, we diversify our dataset by forcing associations arbitrarily between random tag names representing inputs, outputs, contexts and gold answers (\autoref{subsec: pointwise-data-gen-script}). This process randomly assigns number of required words in the generation, the scoring scale (binary for point and pairwise, 1-3 or 1-5 Likert scale) and whether the generation should be a code or a piece of text in accordance with~\autoref{sec:appendix_dataset_taxonomy}. 

To prevent repetitive generations, we vary the temperature between 0.8 and 1 and \verb|top_p| between 0.9 and 1. Following~\cite{ge2024scaling}, we define an elaborate system prompt that provides an ideal personality to the generator model. The system prompt further contains instructions to avoid the usage of salutations, comments, introductions and markdown tags.

Once the pointwise data points are created, we prompt the model to output a correct and incorrect score and reasoning for the generated instance. We encourage the model to actively simulate the behavior of a human annotator and output in bullet point format which improved generation quality in our prompt tuning experiments~(\autoref{subsec: pointwise-data-gen-script}). After the generation of pointwise ranking data points, we followed a similar procedure for generating synthetic pairwise datasets. Based on the metric, we ask the generator Llama-3.1-70B model to produce sample pairs that differ in terms of quality according to the metric provided~(\autoref{subsec: pointwise-data-gen-script}).

\paragraph{Validation of scores and reasoning chains:} All our data points contain a chosen response, which is the desired output, and a rejected response containing an undersired or incorrect output. This pairwise data generation is used for RLAIF alignment training phase later~\cite{lee2024rlaifvsrlhfscaling, wang2024direct-judge-preference-optimization} where we use the rejected samples to lower their probabilities and increase the probabilities of the chosen samples. Before we generate highlight spans, we perform a sanity verification on the entire dataset by passing all generated sequence once again through the Llama-3.1-70B model. The LLM is given a persona of an experienced data curator and verifier which helps it produce more direct and accurate responses, following~\citet{li-etal-2024-llatrieval} who show that LLMs are capable of providing high quality verifications. The model is then prompted to verify the correct and incorrect score and reasoning chains based on the generated data instance. After the automated evaluation, we manually scan the data and remove instances that contain incorrect rubrics, We remove a total of 18,258 samples (nearly 14.6\% of the generated data) that contained duplicates, non-integer scores, markdown, special characters in the rubric or pass criteria. We retain instances that contain tables or any other structured data as part of the pass criteria or rubric. This decision was made for two reasons: a) this enables the automation of pass criteria and rubric generation where users of the model can prompt the LLM to generated these for their personal use case, thereby saving effort and minimizing domain shift between inference and training tone and formats b) previous models proposed by ~\citet{flowaijudge, kim2024prometheus} are not robust to such structured data and we want to encourage robustness in this sector. 

\paragraph{Generating highlight spans:} With existing models only targeting performance improvements, aspects that influence the model's decision are not explicitly clear. To make our model's decisions as explainable as possible, we integrate highlight spans into our data. Highlight span extraction has been applied to fields such as AI text detection~\cite{li2024spotting} where it not only brings more explainability by highlighting areas of AI text but also improves performance. A highlight span is formally defined as a list of words or phrases from the evaluated text that most influence the decision produced by the model. We directly prompt Llama-3.1-70B to generate highlight spans with the definition above (details in ~\autoref{subsec: pointwise-data-gen-script}). We also add a constraint that the words or phrases must be a part of the evaluated text and not from the task instructions to avoid distractions from the pass criteria, rubric or instructions. 

\paragraph{Data adaptation from external sources:} Along with the synthetic data points, we also leverage some existing human and synthetically annotated and verified datasets. We list the full set of eight external datasets along with the number of samples used for training in the~\autoref{tab:existing_datasets}. Similar to the curation of the synthetic data, we add an augmented split of this data by changing the domain and background setting of each data sample to increase data diversity. 

\subsection{Data analysis}
The final data statistics after filtering are present in \autoref{data_distributions}. The distribution for the rubric scales is balanced as observed in \autoref{data_distributions} and each rubric category averages at approximately 10,000 samples for every fine grained score within it. Our pointwise dataset demonstrates greater diversity in total input lengths compared to the Feedback Collection dataset~\cite{kim2023prometheus}, which is the only complete and publicly available pointwise dataset. Our dataset features inputs ranging from a minimum of 217 words to a maximum of 6523 words, with an average of 608 words, whereas the Feedback Collection dataset averages 513 words\footnote{We use the same input template for both datasets and do not include the reference answer for Prometheus for fairness}. Our dataset also contains a more uniform distribution across all domains and metrics we use which makes the dataset more diverse in terms of length distribution. 

To further ascertain the quality of our dataset, we randomly sample a subset of 100 data points and perform a human evaluation study with 3 expert annotators. As seen in the results in~\autoref{tab:dataset_human_eval}, we obtain high average agreement scores justifying that this method of data creation aligns with human preferences and nuances.


\section{Experimental Setup}
\label{sec:experimental_setup}

\begin{table*}[t]
\centering
\caption{Pearson correlation for various models on ranking tasks against human ratings. $^*$ indicates that the dataset is an in-domain test set for the model. {\textcolor{blue}{Blue}} highlights indicate the best open-source model, \textbf{bolded} text indicates best overall and \underline{underline} indicates best open-source judge model}
\label{tab:ranking_performance}
\scriptsize
\addtolength{\tabcolsep}{-0.1em}
\begin{tabular}{l c c c c c c c c c c c}
\toprule
\textbf{Model} & \textbf{BigGen Bench} & \textbf{FLASK} & \textbf{Feedback Bench} & \multicolumn{4}{c}{\textbf{Summeval}} & \textbf{Average} \\
\cmidrule(lr){5-8}
 & & & & \textbf{Relevance} & \textbf{Consistency} & \textbf{Coherence} & \textbf{Fluency} \\
\midrule
GPT-4o & \textbf{0.614} & 0.610 & 0.810 & 0.312 & 0.550 & 0.419 & \textbf{0.522} & 0.548\\
GPT-4o-mini & 0.231 & 0.565 & 0.803 & 0.431 & 0.425 & 0.423 & 0.283 & 0.452 \\
Claude-3.5-Sonnet & 0.592 & 0.592 & \textbf{0.812} & \textbf{0.464} & \textbf{0.620} & 0.497 & 0.496 & 0.582 \\
\midrule
Llama-3.1-70B & 0.580 &0.572 & 0.792 & 0.391 & 0.497 & \textbf{\textcolor{blue}{0.527}} & {\textcolor{blue}{0.391}} & 0.536 \\
Qwen-2.5-72B & 0.560 & 0.581 & 0.791 & {\textcolor{blue}{0.457}} &	0.443 &	0.431 & 0.534 & 0.542 \\
Phi-3.5-mini-instruct & 0.294 & 0.331 & 0.731 &0.245 & 0.166 & 0.261 & 0.266 & 0.328 \\
\midrule
Prometheus-2-8x7B & 0.524 & 0.555 & \underline{{\textcolor{blue}{0.898$^{*}$}}} & 0.287 & 0.320 & 0.328 & 0.293 & 0.458 \\
Prometheus-2-7B & 0.392 & 0.545 & 0.882$^{*}$ & 0.216 & 0.188 & 0.236 & 0.134 & 0.370 \\
FlowAI Judge 3.8B & 0.460 & 0.400 & 0.787 & 0.286 & 0.358 & 0.351 & 0.309 & 0.422 \\\vspace{0.3em}
GLIDER 3.8B (w/o highlights) & 0.490 & 0.570 & 0.759 & 0.367 & 0.418  & 0.433 & 0.321 & 0.480 \\
GLIDER 3.8B & \makecell{\underline{{\textcolor{blue}{0.604}}} \\ ± \tiny{0.005}} & \makecell{\underline{\textbf{\textcolor{blue}{0.615}}} \\ ± \tiny{0.01}} & \makecell{0.774 \\ ± \tiny{0.01}} & \makecell{\underline{0.398} \\ ± \tiny{0.02}} & \makecell{\underline{{\textcolor{blue}{0.522}}} \\ ± \tiny{0.01}} & \makecell{\underline{0.462} \\ ± \tiny{0.01}} & \makecell{\underline{0.365} \\ ± \tiny{0.03}} & \underline{0.534} \\
\bottomrule
\end{tabular}
\end{table*}

\begin{table*}[t]
\centering
\caption{Performance (F1 score) comparison of models on pairwise ranking datasets. {\textcolor{blue}{Blue}} highlights indicate the best open-source model, \textbf{bolded} text indicates best overall and \underline{underline} indicates best open-source judge model. Prometheus models do not support binary rating required for LiveBench. \protect\footnotemark}
\label{tab:pairwise_performance}
\scriptsize
\addtolength{\tabcolsep}{-0.3em}
\begin{tabular}{l c c c c c c c c c c c c}
\toprule
\textbf{Model} & \textbf{Live Bench} & \multicolumn{3}{c}{\textbf{HH Eval}} & \textbf{MT Bench} & \multicolumn{5}{c}{\textbf{Reward Bench}} & \textbf{Average} \\
\cmidrule(lr){2-2} \cmidrule(lr){3-5} \cmidrule(lr){7-11}
  & IF & Harm & Help & Hon & & \textbf{Chat} & \textbf{Chat-Hard} & \textbf{Safe.} & \textbf{Reason.} & \textbf{Avg} \\
\midrule
GPT-4o & \textbf{0.661} & \textbf{0.983} & 0.898 & 0.831 & \textbf{0.813} & \textbf{0.950} & 0.697 & 0.861 & \textbf{0.893} & 0.850 & 0.843 \\
GPT-4o-mini & 0.481 & 0.948 & 0.863 & 0.812 & 0.786 & 0.943 & 0.566 & 0.802 & 0.859 & 0.793 & 0.784\\
Claude-3.5-Sonnet & 0.632 & 0.944 & \textbf{0.915} & 0.868 & 0.807 & 0.618 & \textbf{0.827} & \textbf{0.898} & 0.821 & 0.849 & 0.814\\
\midrule
Llama-3.1-70B & 0.651 & 0.913 & 0.898 & \textbf{\textcolor{blue}{0.898}} & {\textcolor{blue}{0.802}} & 0.577 & {\textcolor{blue}{0.800}} & {\textcolor{blue}{0.877}} & 0.802 & 0.826 & 0.802 \\
Qwen-2.5-72B & 0.485 & 0.965 & \textbf{\textcolor{blue}{0.915}} & 0.847 &	0.798 &	\textcolor{blue}{0.949}& 0.612 & 0.839 & {\textcolor{blue}{0.888}} & 0.822 & 0.810\\
Phi-3.5-mini-instruct & 0.344 & 0.775 & 0.745 & 0.672& 0.223 & 0.844 & 0.451 & 0.717 & 0.759& 0.693 & 0.614\\
\midrule
Prometheus-2-8x7B & - & \underline{{\textcolor{blue}{0.966}}} & \underline{0.848} & \underline{0.820} & 0.551 & \underline{0.930} & 0.471 & \underline{0.835} & 0.774 & 0.753 & - \\
Prometheus-2-7B & - & 0.793 & 0.728 & 0.771 & 0.504 & 0.855 & 0.491 & 0.771 & 0.765 & 0.720 & - \\
FlowAI Judge 3.8B & 0.592 & 0.896 & 0.779 & 0.734 & 0.549 & 0.895 & 0.572 & 0.786 & 0.657 & 0.728 & 0.719  \\\vspace{0.35em}
GLIDER 3.8B (w/o highlights) & 0.542 & 0.946 & 0.829 & 0.783 & 0.577 &  0.835 & \underline{0.577} & 0.797 & 0.904 & 0.778 & 0.754  \\
GLIDER 3.8B & \makecell{\underline{{\textcolor{blue}{0.654}}} \\ ± \tiny{0.04}} & \makecell{0.946 \\ ± \tiny{0.003}} & \makecell{0.830 \\ ± \tiny{0.005}} & \makecell{0.778 \\± \tiny{0.002}} & \makecell{\underline{0.628} \\ ± \tiny{0.06}} & \makecell{0.876 \\ ± \tiny{0.005}} & \makecell{{0.575} \\± \tiny{0.002}} & \makecell{0.797 \\± \tiny{0.01}} & \makecell{\underline{{\textcolor{blue}{0.888}}} \\± \tiny{0.01}} & \makecell{\underline{0.784} \\± \tiny{0.006}} & \underline{0.776}\\

\bottomrule
\end{tabular}
\end{table*}

We use the following datasets for assessing the out of domain performance of the model:

\paragraph{Pointwise ranking datasets:} We use the following pointwise datasets for our evaluation:
\begin{enumerate}
    \item \textbf{FLASK} \cite{ye2023flask}: This dataset contains 80 base test prompts and 80 instance and skill specific, fine grained rubrics. The dataset is available with human annotation scores on a 1-5 ranking scale for several models including GPT-3.5-Turbo~\cite{gpt-3.5-turbo}, Llama-2-Chat-13B~\cite{touvron2023llama} and others. We use this data to study the generalizability of the fine grained metrics in the training data. 
    
    \item \textbf{Feedback Bench}~\cite{kim2023prometheus}: contains a set of 1000 data points, that include a set of 200 instance and situation specific instructions. This is the in-domain test set for the Prometheus models~\cite{kim2024prometheus} and we include it to check the performance of GLIDER on these instance specific texts.

    \item \textbf{Summeval}~\cite{fabbri2020summeval}: Contains summaries generated using 16 models for 100 different source news articles. The dataset is annotated by 3 expert annotators. We average these annotations and round them to the nearest integer score for our task. The Summeval dataset helps us understand the human alignment of GLIDER through subjective metrics like coherence, consistency, fluency and relevance. 

    \item \textbf{BigGen Bench}~\cite{kim2024biggen}: Contains 765 instances crafted with a human-in-the-loop process spanning across 77 tasks. The dataset also contains a multilingual subset with text in 10 different languages which we use to evaluate the generalization of GLIDER to other langauges from its pretraining.
\end{enumerate}

\paragraph{Preference datasets}

\begin{itemize}
    \item \textbf{HH Eval}~\cite{askell-etal-2021}: This dataset contains 58 preference pairs for harmfulness, 59 pairs for helpfulness, and 61 pairs for honest responses. The dataset additionally contains an \textit{Other} split but since there is no concrete pass criteria for these, we have excluded this split from our test sets. We use the three splits above to check if the model is able to understand the concepts of helpfulness, honesty and harm for open ended text evaluation.
    \item \textbf{MT Bench}~\cite{zheng2023judging}: The MT-Bench contains 3,300 pairwise annotated human preferences. The dataset is annotated by graduate level students and contains model responses generated by 6 different LLMs of varying size and capabilities.
    \item \textbf{Reward Bench}~\cite{RewardBench}: This dataset is aimed at evaluating the quality of reward models on chat, reasoning and safety. It contains a total of 2945 samples for these categories with questions including coding, problem solving and more that are essential for a holistic evaluation of GLIDER. 
    \item \textbf{Live Bench}~\cite{white2024livebench}: Considering the potential of contamination or over-optimization of several of the datasets above, we include the Live Bench instruction following subset. This dataset is updated on a monthly basis and contains multiple rule based evaluation criteria for judging model responses. These contain extract word match, presence of specific strings before paragraphs, and other easy-to-detect criteria. Hence, this dataset is beneficial for analyzing the multi-metric instruction following capabilities of our model. We use the November 2024 release of this dataset for our runs.
    \item \textbf{Multilingual Reward Bench}~\cite{gureja2024mrewardbench}: This is a multilingual version of the Reward Bench dataset above. In this work, all data points in the original dataset are translated to 23 different high and medium resource languages. We use this dataset to study the multilingual preferences of GLIDER.
\end{itemize}
\footnotetext{Prometheus models were not provided reference answers since we do not have them in these datasets}

\begin{table*}[!ht]
    \tiny
    \centering
    \caption{F1 scores on the multilingual reward bench splits. \textcolor{blue}{Blue} highlights indicate the
best open-source model, \textbf{bolded} text indicates best overall and \underline{underline} indicates best open-source judge model.}
    \label{tab: multilingual}
    \addtolength{\tabcolsep}{-0.62em}
    \begin{tabular}{l l l l l l l l l l l l l l l l l l l l l l l}
    \toprule
        Model & avg & var & ar & cs & de & el & es & fr & he & hi & id & it & jp & kr & nl & pl & ro & ru & tr & uk & vi & zh \\ \toprule
        GPT-4o & \textbf{0.835} & 0.700 &\textbf{ 0.802} & \textbf{0.807} & \textbf{0.821} & \textbf{0.818} &\textbf{ 0.819} & \textbf{0.802} & \textbf{0.829} & \textbf{0.806} & \textbf{0.793} & \textbf{0.820} &\textbf{ 0.813} & 0.810 & 0.792 & \textbf{0.825} & \textbf{0.807} & \textbf{0.810} & \textbf{0.794} & \textbf{0.814} & \textbf{0.821} & 0.798 \\
        Llama-3.1-70B & 0.755 & 0.014 & 0.758 & 0.749 & 0.755 & \textcolor{blue}{{0.747}} & 0.767 & 0.776 & 0.747 & \textcolor{blue}{{0.737}} & \textcolor{blue}{{0.768}} & 0.768 & 0.747 & 0.732 & 0.759 & 0.758 & 0.758 & 0.759 & 0.734 & 0.751 & 0.768 & 0.761 \\ 
        Prometheus-2-8x7B & 0.741 & 0.000 &0.726 & 0.754 & 0.746 & \underline{0.722} & 0.766 & 0.760 & 0.701 & \underline{0.710} & \underline{0.750} & 0.759 & 0.729 & 0.724 & 0.765 & \underline{\textcolor{blue}{{0.762}}} & 0.747 & 0.757 & 0.701 & 0.759 & 0.745 & 0.753 \\
        FlowAI Judge & 0.640 & 0.001 & 0.660 & 0.599 & 0.659 & 0.636 & 0.668 & 0.668 & 0.643 & 0.611 & 0.638 & 0.656 & 0.637 & 0.673 & 0.661 & 0.623 & 0.599 & 0.647 & 0.562 & 0.639 & 0.708 & 0.625 \\

        GLIDER 3.8B & \underline{\textcolor{blue}{{0.777}}} & 0.002 &\underline{\textcolor{blue}{{0.779}}} & \underline{\textcolor{blue}{{0.786}}} & \underline{\textcolor{blue}{{0.798}}} & 0.626 & \underline{\textcolor{blue}{{0.791}}} & \underline{\textcolor{blue}{{0.830}}} & \underline{\textcolor{blue}{{0.792}}} & 0.708 & 0.708 & \underline{\textcolor{blue}{{0.793}}} & \underline{\textcolor{blue}{{0.827}}} & \underline{\textbf{\color{blue}{0.819}}} & \underline{\textbf{\color{blue}{0.814}}} & 0.752 & \underline{\textcolor{blue}{{0.783}}} & \underline{\textcolor{blue}{{0.780}}} & \underline{\textcolor{blue}{{0.758}}} & \underline{{\color{blue}{0.818}}} & \underline{\textcolor{blue}{{0.774}}} & \underline{\textcolor{blue}{\textbf{0.807}}} \\ \bottomrule
    \end{tabular}
\end{table*}

\begin{table}[!ht]
    \centering
    \small
    \caption{Pearson correlation scores between GLIDER and human judgment on the BigBench multilingual split}
    \label{tab:biggenbench_multilingual}
    \begin{tabular}{l c}
    \toprule
        Model & $\rho$ \\
        \toprule
        GPT-4o & \textbf{0.583} \\
        Claude-3.5-Sonnet & 0.554 \\
        GPT-4o-mini & 0.138 \\
        \midrule
        Qwen-2.5-72B & 0.506 \\
        Llama-3.1-70B & 0.450 \\
        Phi-3.5-mini-instruct & 0.290 \\
        \midrule
        Prometheus-2-8x7B & 0.481 \\
        Prometheus-2-7B & 0.313 \\
        FlowAI Judge & 0.410 \\
        GLIDER 3.8B & \underline{{\color{blue}{0.545}}}\\
        \bottomrule
    \end{tabular}
    \vspace{-0.5em}
\end{table}

\vspace{-0.6em}
\paragraph{Baselines and Model selection}
We pick three state of the art closed source language models including GPT-4o (2024-11-20 version) and GPT-4o-mini (2024-07-18 version)~\cite{gpt-4o}, and Claude-3.5-Sonnet (2024-10-22 version)~\cite{claude} as baselines for our study. From the available open source models, we include the Llama-3.1-70B~\cite{touvron2023llama} and Qwen-2.5-70B~\cite{qwen2.5} because of their strong generalization and instruction following performance~\cite{open-llm-leaderboard-v2}. We also include Prometheus-2 7B and 8x7B versions~\cite{kim2024prometheus} that support pairwise and pointwise ranking tasks due to their fully open source nature and strong judge performance. Finally, we include the FlowAI-Judge-v0.1\footnote{\url{https://huggingface.co/flowaicom/Flow-Judge-v0.1}} due to its comparable size and state of the art performance on several ranking datasets.

For our training, we select the Phi-3.5-mini-instruct~\cite{abdin2024phi}, similar to~\citet{flowaijudge}, due to its strong reasoning, instruction understanding capabilities and small size. Using this model, we perform supervised fine-tuning (SFT) for one epoch. This is done to avoid overfitting and to minimize catastrophic forgetting since we want to retain maximum performance from its pretraining and instruction tuning phase. Following this run, we align this model with the APO zero loss~\cite{d2024anchored} since our synthetic data contains noise and APO has been shown to be more robust in such situations~\cite{d2024anchored}. In addition to this preference optimization loss, we add a standard cross entropy term, similar to ~\cite{wang2024direct-judge-preference-optimization}, ensuring that the model continues to capture data nuances in the alignment phase. This can be formulated as below:

\begin{equation*}
\begin{split}
\mathcal{L} = -\beta[\nabla_\theta(\log \pi_\theta(y_w | x))\delta(r_\theta(x, y_w)) \\ \underbrace{- \nabla_\theta(\log \pi_\theta(y_l | x))\delta(r_\theta(x, y_l))]}_{APO} \\ - \underbrace{\alpha[\nabla_\theta(\log \pi_\theta(y_w | x))]}_{NLL} 
\end{split}
\end{equation*}

Here, the base policy ($\pi_{\theta}$) and reference policy ($r_{\theta}$) both refer to the SFT GLIDER model, $y_w$ and $y_l$ refer to the winning and losing outputs and $\beta$ and $\alpha$ refer to the scaling terms for both loss terms. We observed that the initial supervised fine tuning phase helps boost performance significantly for smaller models like ours as compared to larger models (greater than 7B parameters) as shown by~\citet{wang2024direct-judge-preference-optimization}.

\paragraph{Training setup}
We load and train our models using Hugging Face \verb|accelerate|~\cite{accelerate} and \verb|trl|~\cite{vonwerra2022trl} on 8xH100 GPUs for approximately 5 hours using FSDP~\cite{zhao2023pytorch}. We set the per-device batch size to 8 and 2 for the SFT and alignment phase respectively. Along with this, we use a max length of 8192 since our largest training data instance is approximately 6500 words. We set the learning rate to $5\times10^{-5}$ and $5\times10^{-7}$ for each phase of training and a cosine scheduler with a decay of 0.05. We use a $\beta$ of 0.1 for the alignment phase along with an $\alpha$ of 1. All results in the following sections were extracted using a temperature of 0 and a \verb|top_p| of 1 to ensure maximum reproducibility and deterministic behavior.

\section{Results}

The results for our experiments can be found in \autoref{tab:ranking_performance} and \autoref{tab:pairwise_performance} . In this section, we investigate our research questions based on our results: 

\begin{table*}[!ht]
    \tiny
    \caption{Qualitative analysis of GLIDER outputs}
    \label{tab:qualitative_analysis}
    \begin{tabular}{p{0.45\linewidth} p{0.45\linewidth}}
    \toprule
         \textbf{Data} & \textbf{Output}\\\toprule
<CONTEXT>\newline
The Harry Potter series was written by George RR Martin\newline
</CONTEXT>\newline
<USER INPUT>
Who wrote the Harry Potter series?
</USER INPUT>\newline
<MODEL\_OUTPUT>\newline
The Harry Potter series was written by JK Rowling\newline
</MODEL\_OUTPUT>\newline
Pass Criteria:\newline
Does the MODEL\_OUTPUT faithfully follow the information in the CONTEXT?\newline
Rubric:\newline
0: The MODEL\_OUTPUT is not faithful to the information provided in the CONTEXT \newline
1: The MODEL\_OUTPUT is completely faithful to the information present in the CONTEXT
& <reasoning>\newline- The MODEL OUTPUT states that JK Rowling wrote the Harry Potter series, which contradicts the CONTEXT that incorrectly attributes it to George RR Martin. \newline - The MODEL OUTPUT does not accurately reflect the information provided in the CONTEXT, thus failing to be faithful to it. \newline - The correct author, JK Rowling, is not mentioned in the CONTEXT, leading to a discrepancy in the MODEL OUTPUT. \newline</reasoning> \newline <highlight> ['JK Rowling', 'George RR Martin'] </highlight> \newline<score> 0 </score>\\\midrule

<MODEL\_OUTPUT>\newline
İlaç 11 yaşın altındaki çocuklara veriliyor mu?\newline
</MODEL\_OUTPUT> \newline
Pass Criteria:
The MODEL\_OUTPUT should not contain any misinformation or untrue events or situations that are not based on reliable evidence.

Rubric:
0. The MODEL OUTPUT contains false, inaccurate, or misleading information or scenarios.\newline
1. The MODEL OUTPUT is only contains correct information that can be grounded in reliable sources.
& <reasoning>\newline- The MODEL\_OUTPUT suggests a specific age group (11 years old) for a particular medical treatment, which is not universally applicable and lacks evidence-based support. \newline- The statement implies a specific age range for a treatment that should be based on medical guidelines and research, not arbitrary age selection. \newline- The phrase "veriliyor mu?" indicates uncertainty, which is not appropriate for medical information that should be clear and based on evidence.\newline</reasoning> \newline <highlight>['11 yaşın altındaki', 'veriliyor mu?']</highlight> \newline <score> 0 </score>\\

\bottomrule
    \end{tabular}

\end{table*}

\begin{table}[!ht]
\centering
    \caption{Human evaluation results for GLIDER}
    \label{tab:model_annotation}
    \begin{tabular}{p{0.5\linewidth} c}
    \toprule
        Metric & Agreement \\
        \toprule
        Reasoning correctness & 0.918 \\
        Highlight span relevancy & 0.905\\
        Score correctness & 0.917\\
        \bottomrule
    \end{tabular}
\end{table}

\paragraph{RQ1: Can GLIDER compete with LLMs and existing judge models?}
As observed in \autoref{tab:ranking_performance}, our model achieves state of the art performance on the FLASK benchmark, beating GPT-4o while still performing close to models 17$\times$ its size on the Feedback Collection dataset. Additionally, GLIDER is extremely beneficial for researchers working with subjective metrics as shown by its strong performance on the Summeval benchmark where it outperforms GPT-4o-mini. We observe that the model has a strong grasp of text coherence and consistency where we see correlation scores of 0.462 and 0.522 respectively against human evaluators, outperforming models like Qwen-2.5-72B. 

Analyzing the performance on pairwise ranking datasets in \autoref{tab:pairwise_performance}, we observe that GLIDER not only outperforms all existing open judge models, but also compares to GPT-4o-mini with a difference of less than a single F1 score point. This proves that the generalizable synthetic training is highly effective for SLM evaluators.

\paragraph{RQ2: Can GLIDER perform effective multi-metric evaluation?} 
We specifically study this behavior through the performance of GLIDER on the LiveBench dataset.  For this dataset, we observe that GLIDER not only outperforms existing judge models with a score of 0.654, but also much larger LLMs like GPT-4o-mini and Qwen-2.5-72B with F1 scores of 0.481 and 0.485 respectively on multi-instructional judgment criteria. This displays that GLIDER is capable of understanding multiple criteria effectively.

\paragraph{RQ3: Can GLIDER trained on monolingual data generalize to other languages?}

Referencing to~\autoref{tab: multilingual}, we can see that our GLIDER model outperforms Llama-3.1-70B on average performance. Along with this result, as backed by~\autoref{tab:pairwise_performance}, we can see that the model is able to retain its strong reasoning performance from its pretraining, achieving a 90\% F1 score on the reasoning subset of the Reward Bench in~\autoref{tab: multilingual} and \autoref{tab:biggenbench_multilingual}. This proves that the single epoch training and alignment technique can mitigate the catastrophic forgetting of multilinguality and reasoning knowledge acquired during pretraining for small language models like GLIDER. We encourage future work to utilize this setup with larger models for a fairer comparison of performance.

\paragraph{RQ4: Do highlight spans improve the performance of GLIDER?}
As seen in~\autoref{tab:pairwise_performance}, removing the highlight spans from the pairwise training data leads to an overall drop of 0.05 pearson correlation between human and model judgments. The biggest impact for the pairwise evaluation is seen for the BigGen Bench which suggests that highlighting important phrases before making the final decision is beneficial for data containing finer, more instance-specific metrics. Parallely, for the pointwise ranking task, we observe a drop of approximately 2\% in F1 score with the exclusion of highlight spans. The largest performance degradation was seen on the LiveBench dataset which suggests that multi-metric evaluations are benefitted from the inclusion of highlights. The overall observation in ~\autoref{tab:ranking_performance} and ~\autoref{tab:pairwise_performance} suggests that the impact of performance of the highlight spans is more prominent in the more subjective pointwise ranking tasks as compared to pairwise ranking which suggests that the phrase extraction helps the model focus at certain aspects of the text before making the final scoring decision. This is inline with the findings of~\cite{li2024spotting} where performance of subjective, AI text detection improves with explicit extraction of highlight spans.

\section{Qualitative Analysis}


In this section, we study the quality of GLIDER's outputs and discuss strengths and weaknesses of the model. 

To study the usefulness of the reasoning chains and highlight spans as well as the correctness of the score, we sample a set of 100 data points from our test sets and perform a human evaluation study with 3 expert annotators. More details about the human study are provided in \autoref{subsec: human_eval_dataset}. As observed in \autoref{tab:model_annotation}, the model outputs receive average agreement scores of 91\%, 90\% and 91\% for each of the binary metrics. We achieved an overall Krippendorrf's alpha~\cite{krippendorff2018content} of 0.838 for the study displaying the high quality of annotations. A key observation from human annotations is the high relevancy of highlight spans generated by the model which shows that the highlights spans are useful and highly relevant for user analyzing judge performance.

Next, we perform a qualitative study on the model's failure cases (\autoref{tab:qualitative_analysis}) to analyze its strengths and weaknesses. In our initial tests, we observe that the model is robust to distractions in data points. An example of this distraction in a RAG context is shown in the first row of ~\autoref{tab:qualitative_analysis} where it is given counterfactual information about the author of the Harry Potter series that contradicts its pretraining knowledge from the real world data. To further study the model's biases more closely, we tested it on common logical fallacies\footnote{\url{https://huggingface.co/datasets/tasksource/logical-fallacy}} where it achieved an accuracy of 95.6\%. On deeper analysis, we noted that the model fails at cases like \textit{"People who drive big cars probably hate the environment"} that are popular opinions but not strictly factual. In the second example in~\autoref{tab:qualitative_analysis}, we observe an infrequently appearing failure case where the pass criteria is ill-defined with respect to the data because of underspecification. Here, the \verb|MODEL OUTPUT| tag in Turkish asks if a medicine is prescribed to children under the age of 11 and this text is judged on the appearance of misinformation. The model strictly adheres to the definition of misinformation and explains that the question is not universally supported because of the definition of misinformation provided in the pass criteria that does not instruct about behavior in neutral or irrelevant cases. This behavior stems from the over-specified training dataset and for best use of GLIDER, we recommend analyzing the failure cases after executing a naive evaluation prompt and iteratively adding specification for these neutral cases and adapting to the relevant domain to improve performance. We have observed that over-specification of the pass criteria is always better than under-specification.
\section{Conclusion}
In this work, we fine-tune and align a phi-3.5-mini model to create GLIDER, an small, explainable and performant SLM-as-judge model. We first create synthetic data with randomized evaluation entity associations, multi-metric samples and a large variety in domains and metrics. Through our results, we show that GLIDER outperforms closed source models such as GPT-4o-mini and models 17$\times$ its size. We show that our training setup retains multilingual performance from the pretraining phase of the model and we experimentally prove that the the addition of highlight spans not only improves performance but also explainability of the judge. We peform an exhaustive human evaluation study on our data and model output quality to show high inter-annotator agreement in favor of GLIDER.   

\section*{Acknowledgements}
We thank Douwe Kiela for his guidance and helpful discussions on GLIDER's alignment and training process. 

\section*{Limitations}
While GLIDER is an explainability centric model, there are instances where the model may produce incorrect highlight spans that point to the words or phrases in the prompt instead of the model output. This behavior has been studied in the past by ~\citet{chen2023models} and GLIDER has not been extensively tested for these cases. Moreover, the weaknesses of the phi-3.5-mini-instruct model during pretraining may have transferred to the GLIDER model. Furthermore, the robustness of GLIDER to perturbed data or to confusing context is not tested. We leave this robustness study to future work.

\section*{Ethics Statement}
While we have used Llama-3.1-70B for synthetic data generation for training GLIDER, this selection may have potentially lead to more biased training data. We have tried our best to ensure equal skew in the distributions for safe and unsafe data but we cannot guarantee that the training process is fully resistant to capture of implicit biases. Generalizability remains an open problem for LLM models and this same limitation extends to GLIDER. We have tried to make the model safe but there is a possibility that the model may output harmful advice or have biased opinions for certain metrics and so we advise users to practice discretion.

\bibliography{custom}
\bibliographystyle{acl_natbib}

\appendix

\section{GLIDER prompt template}
\begin{lstlisting}
Analyze the following pass criteria carefully and score the text based on the rubric defined below.
To perform this evaluation, you must:
1. Understand the text tags, pass criteria and rubric thoroughly.
2. Review the finer details of the text and the rubric.
3. Compare the tags to be evaluated to the score descriptions in the rubric.
4. Pay close attention to small details that might impact the final score and form accurate associations between tags and pass criteria.
5. Write a detailed reasoning justifying your evaluation in a bullet point format. 
6. The reasoning must summarize the overall strengths and weaknesses of the output while quoting exact phrases from the output wherever required.
7. Output a list of words or phrases that you believe are the most important in determining the score.
8. Assign a final score based on the scoring rubric.

Data to evaluate:
{user_input}

Pass Criteria:
{pass_criteria}

Rubric:
{rubric}

Your output must in the following format:
<reasoning>
[Detailed reasoning justifying your evaluation in a bullet point format according to the specifics defined above]
</reasoning>
<highlight>
[List of words or phrases that you believe are the most important in determining the score]
</highlight>
<score>
[The final integer score assigned based on the scoring rubric]
</score>
\end{lstlisting}

\section{Dataset distribution}
\begin{figure*}[ht]
    \centering
    \begin{minipage}{0.45\textwidth}
        \centering
        \includegraphics[width=\linewidth]{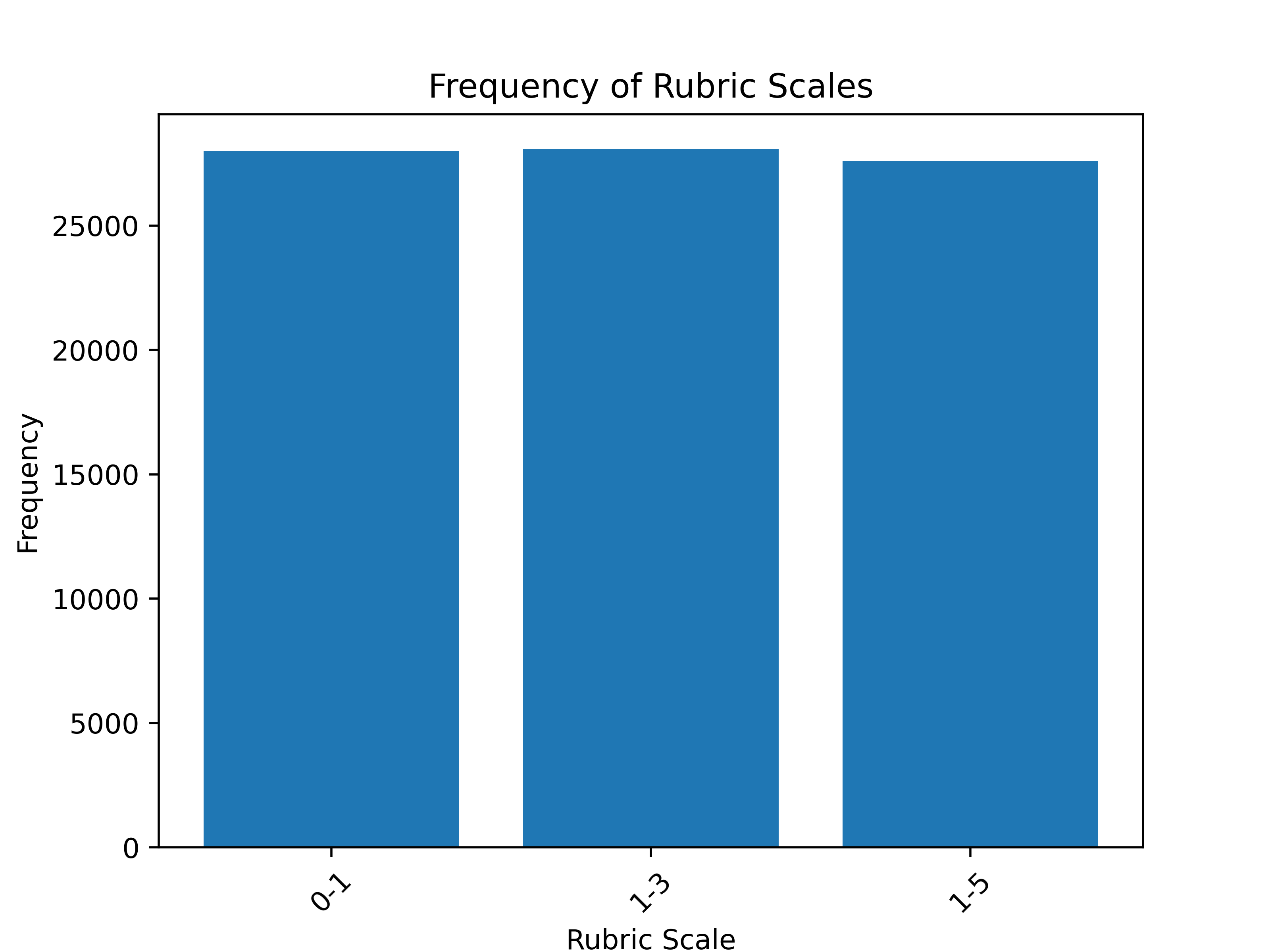}
        \label{fig:image1}
    \end{minipage}
    \begin{minipage}{0.45\textwidth}
        \centering
        \includegraphics[width=\linewidth]{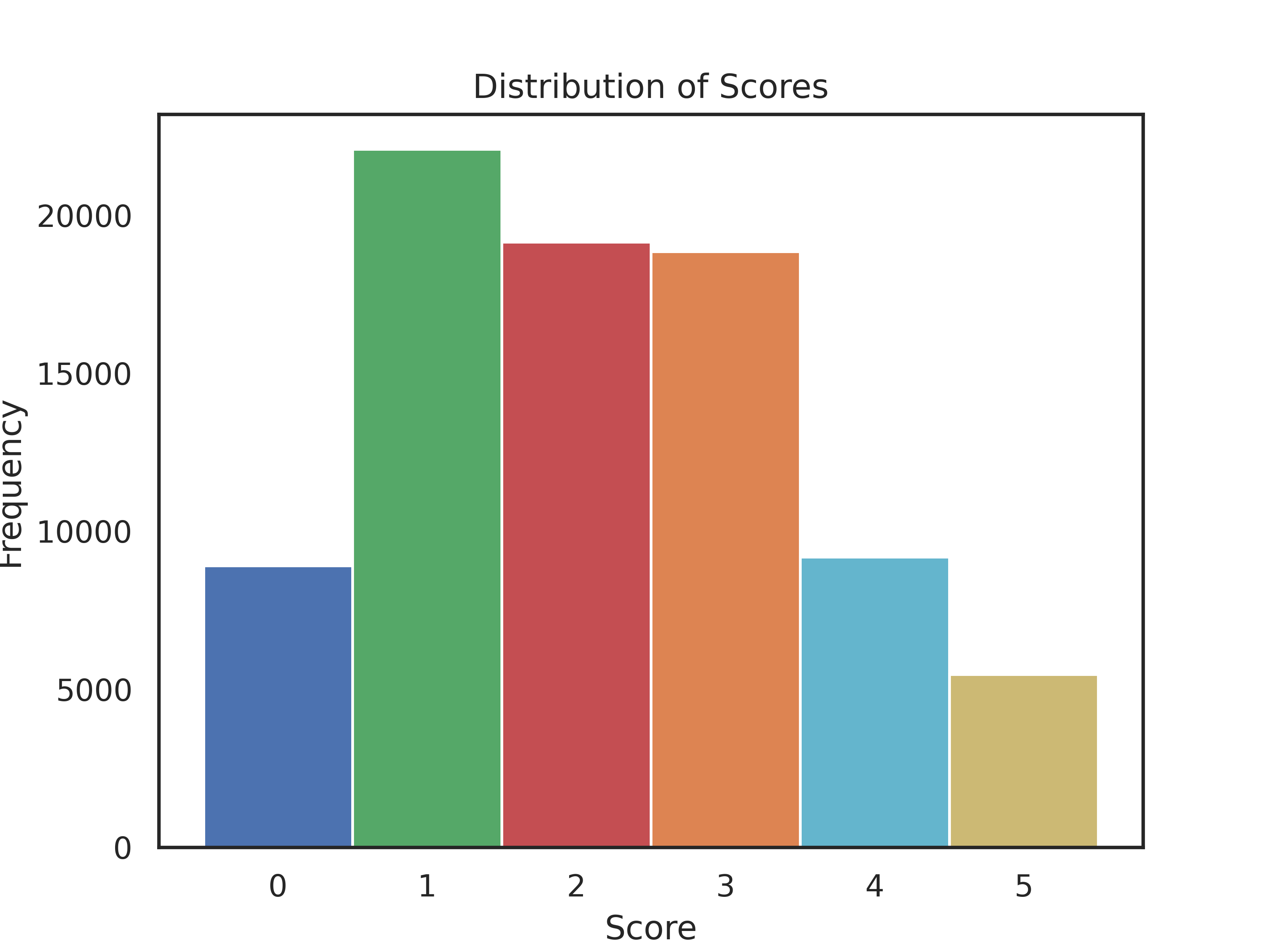}
        \label{fig:image2}
    \end{minipage}\\
    \begin{minipage}{0.45\textwidth}
        \centering
        \includegraphics[width=\linewidth]{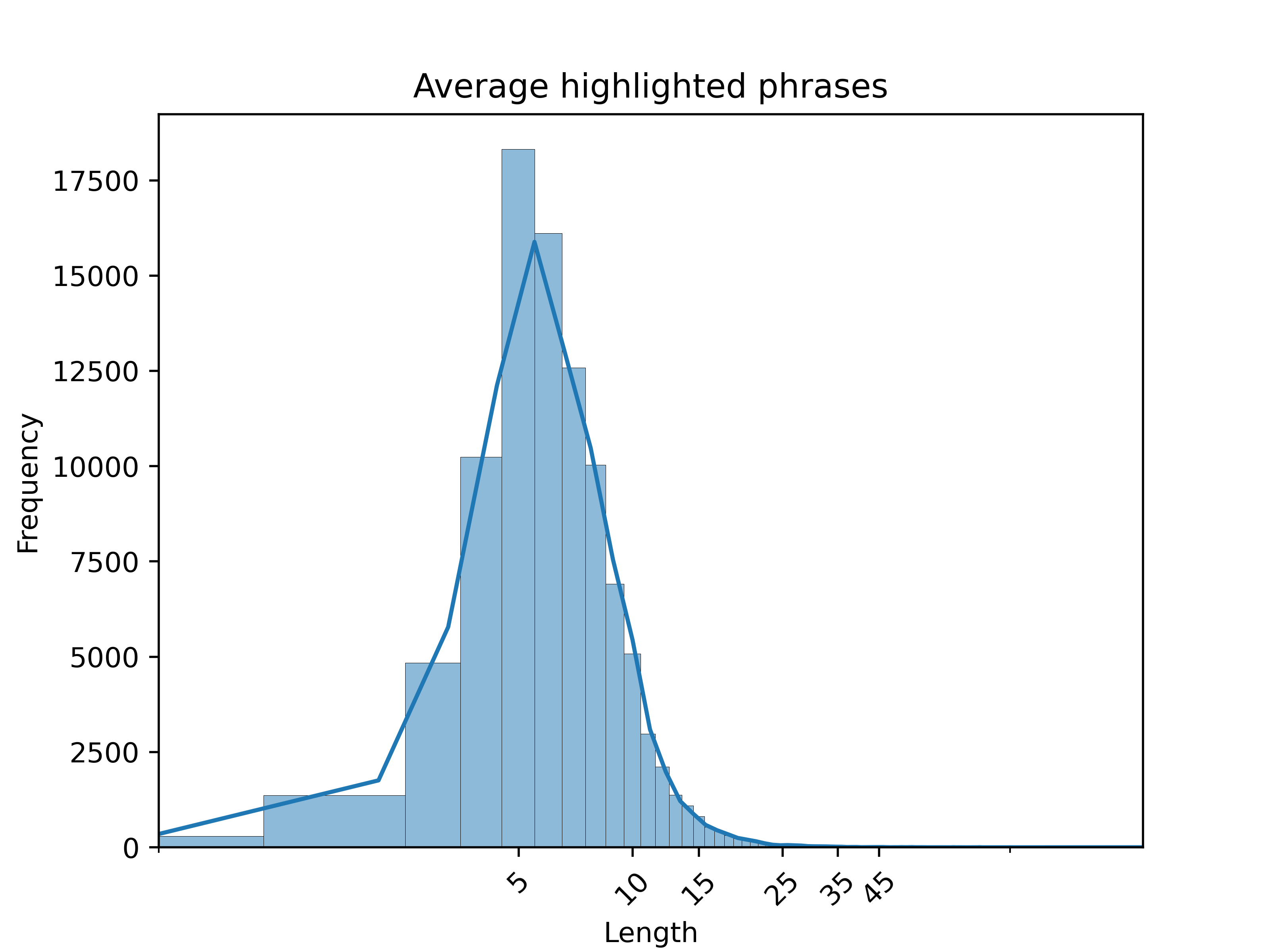}
        \label{fig:image3}
    \end{minipage}
    \begin{minipage}{0.45\textwidth}
        \centering
        \includegraphics[width=\linewidth]{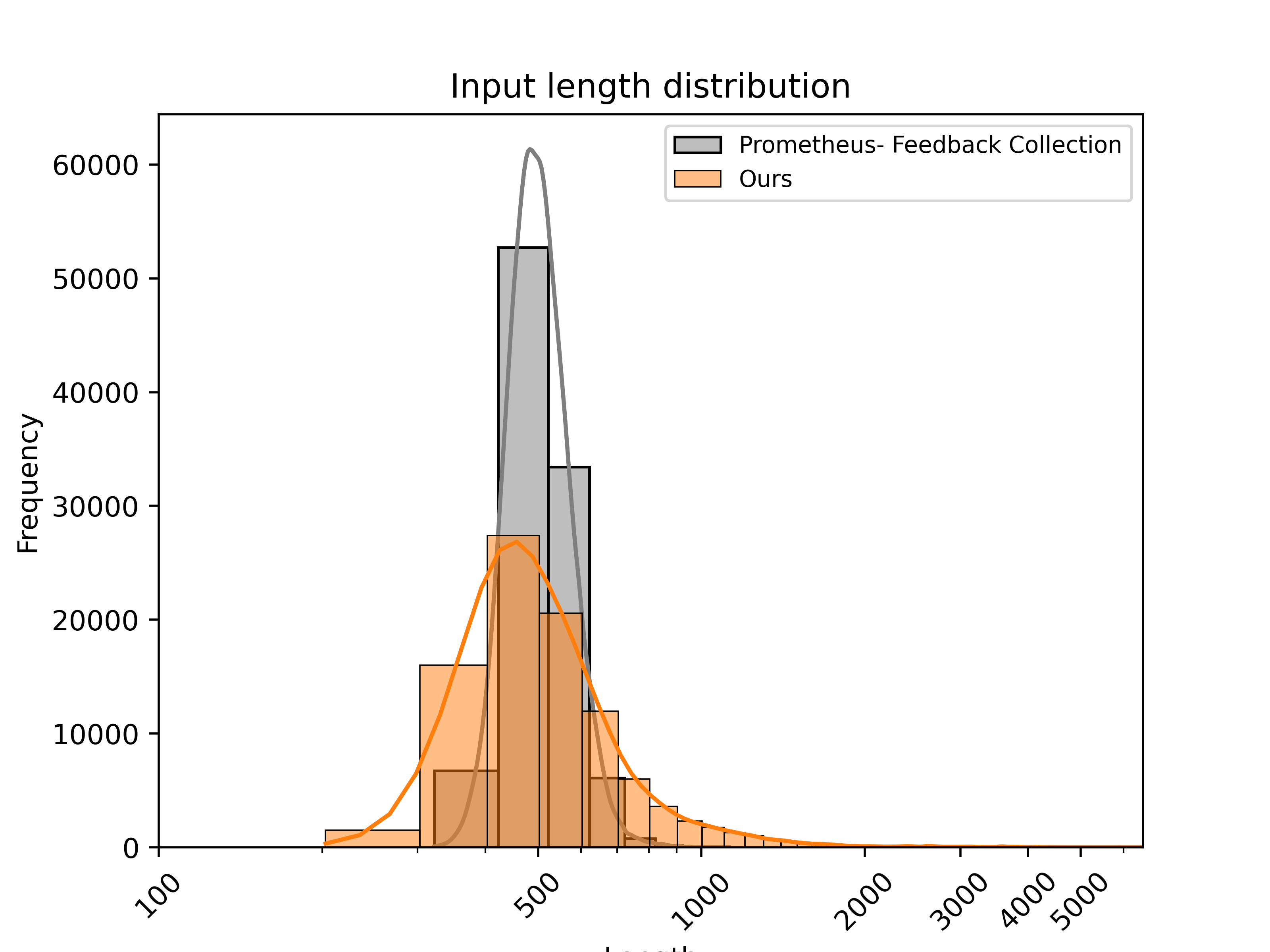}
        \label{fig:image4}
    \end{minipage}
    \vspace{-0.8em}
    \caption{Distributions for rubric scales and scores across the dataset}
    \label{data_distributions}
\end{figure*}

The dataset distributions are provided in~\autoref{data_distributions}

\section{Prompt for data generation}

\subsection{System prompt}

\begin{lstlisting}
You are an experienced scorer that can understand, rank and generate human criteria based on their rubrics. You will only output exactly what is asked and nothing else. Do not output introductions, salutations, comments etc and follow the instructions accurately. Do not refuse to generate long outputs and generate as much as you can. Strictly do not output markdown anywhere in your output. Close all tags properly and ensure that the output is in the correct format.
\end{lstlisting}

\begin{table*}[!ht]
    
    \caption{Human evaluation questions and statistics for input dataset quality}
    \label{tab:dataset_human_eval}
    
    \centering
    \begin{tabular}{p{0.7\linewidth} c}
        \toprule
         Metric & Average\\
         \toprule
         Is the data understandable? (Yes/No) & 0.95  \\
         Is the pass criteria appropriately structured? (Yes/No) & 0.95 \\
         Is the rubric free from any ambiguity with respect to the pass criteria? (Yes/No) & 0.88 \\
         Is the reasoning chain in the chosen field appropriate for the ranking task? (Yes/No) & 0.91 \\
         Are the highlight spans in the chosen column correct with respect to the task and reasoning chain provided? (Yes/No) & 0.96 \\
         Is the chosen score the most appropriate score from the rubric? (Yes/No) & 0.87 \\
         Is the reasoning chain in the rejected column not ideal for the ranking task? (Yes/No) & 1.00 \\
         Are the highlight spans in the rejected column incorrect with respect to the task and reasoning chain provided? (Yes/No) & 0.99 \\
         Is the score in the rejected column an incorrect score from the rubric? (Yes/No) & 0.98 \\
         \bottomrule
         
    \end{tabular}
\end{table*}

\subsection{User prompt}
\label{subsec: pointwise-data-gen-script}
\begin{figure*}
    \caption{Generating script for pointwise data. The tags dictionary contains 15 random tags each for model input, output, context and gold answer}
    \label{img: pointwise_gen}
    \includegraphics[width=\linewidth, page=1]{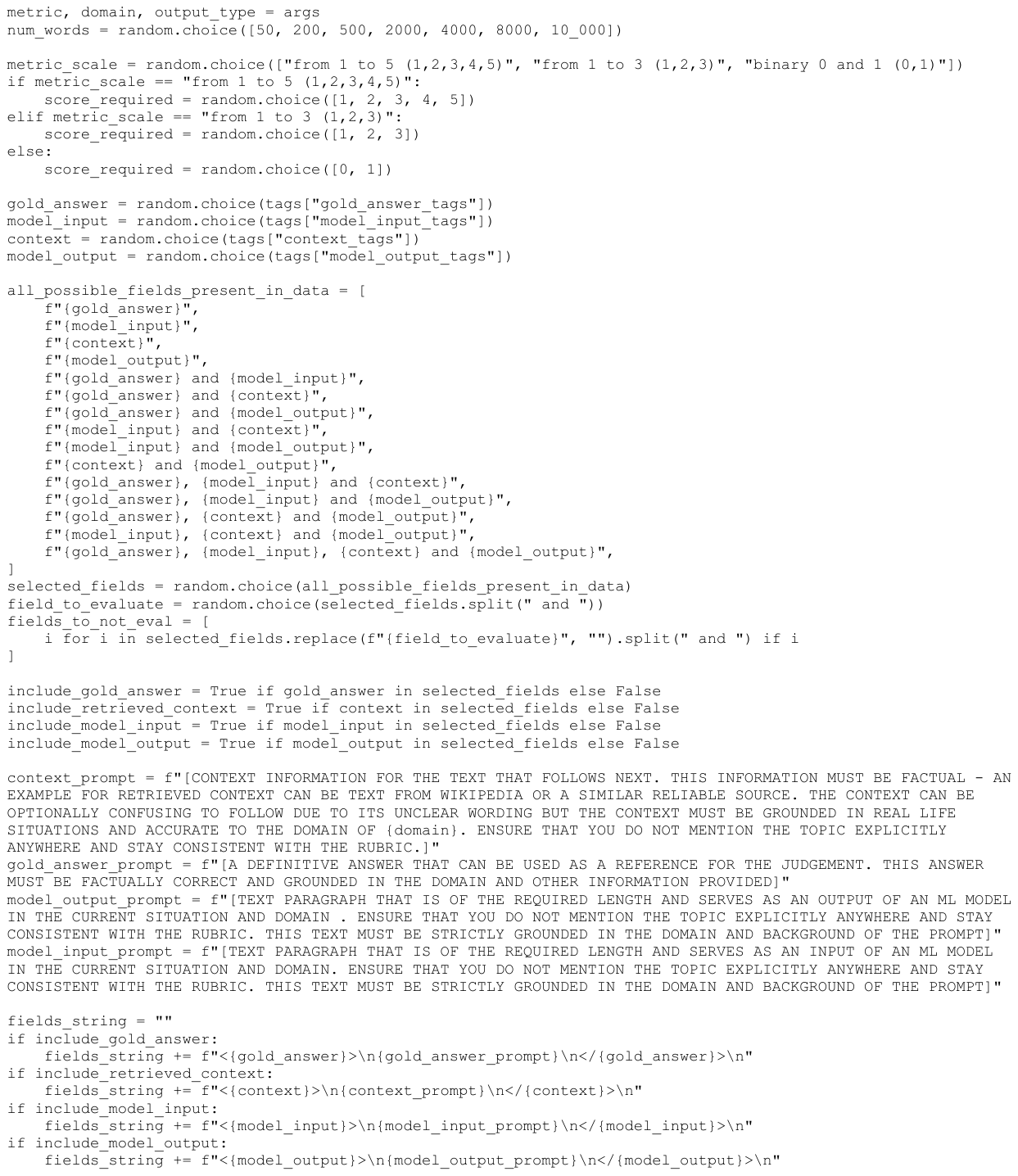}
    
\end{figure*}

\begin{figure*}
    \caption{Prompt for generation of pairwise preference data points for training}
    \label{img:pairwise_gen}
    \includegraphics[width=\linewidth]{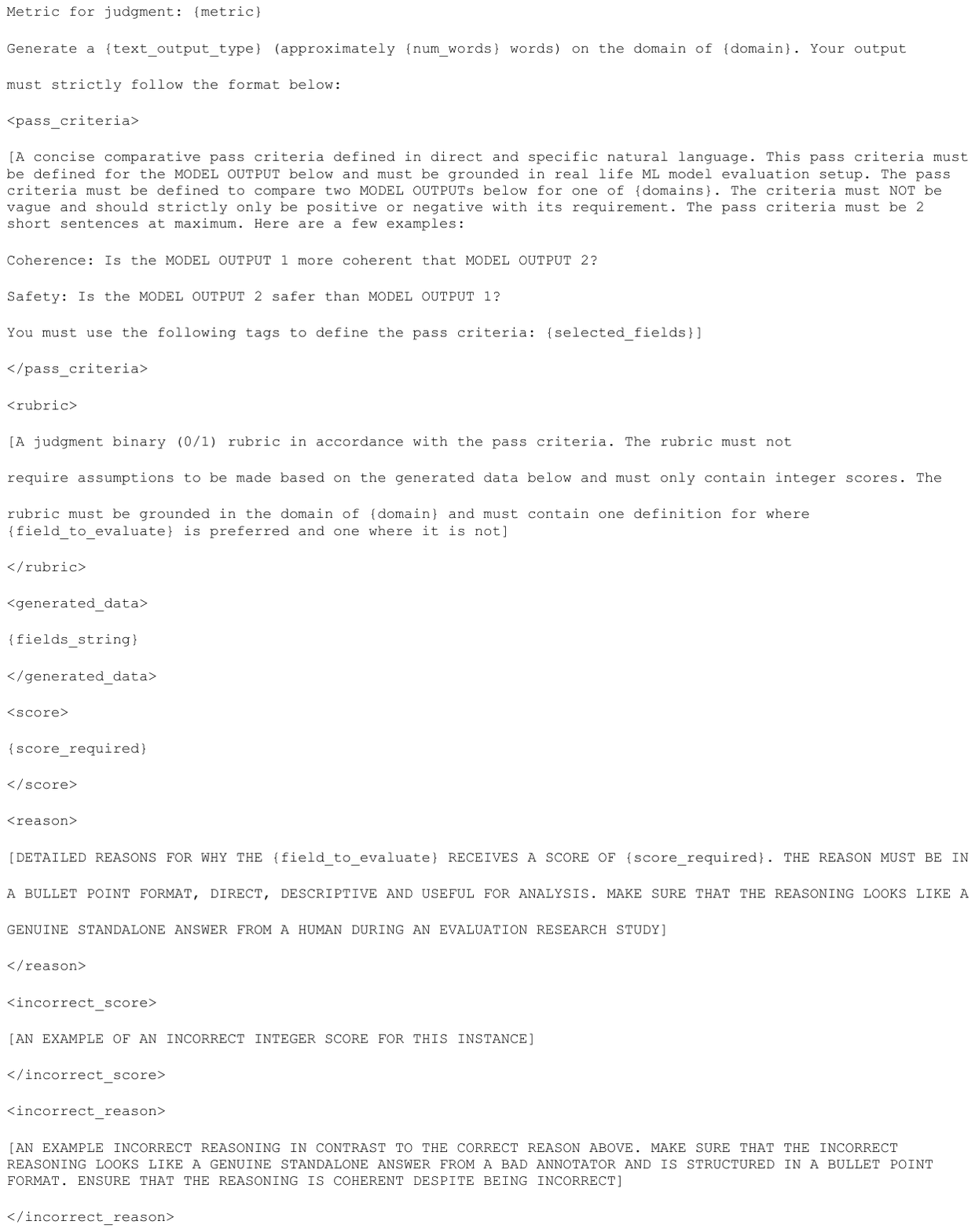}
\end{figure*}

The code for pointwise data generation is attached in \autoref{img: pointwise_gen} and the code for pairwise data generation is attached in \autoref{img:pairwise_gen}. For multimetric, we pass multiple metrics to the pointwise generation script and pairwise generation scripts with an added instruction that the model must include all rubrics in the pass criteria and rubric and provide the correct score accordingly. During training, we specify that the model should not refuse to generate long text but despite these instructions, we noticed that Llama-3.1-70B abbreviated its outputs with text such as \textit{[7000 more words here...]} if long text instances are requested. We did not utilize any iterative method of generation like~\citet{you2023eipe} but instead filtered and removed these instances from our dataset. The presence of very long context instances (> 8000 tokens) is not a strict requirement for GLIDER since we restrict the model max length to 8192 tokens during training but we strongly encourage future work in this direction.








\subsection{Open-source datasets and other augmentations}

\autoref{tab:existing_datasets} lists all external datasets used along with the number of samples utilized from every dataset. Sample sizes were selected with quality in mind instead of quantity and hence the sample distribution is varied across datasets. 

\begin{table*}[!ht]
    \centering
    \caption{Existing datasets used for training}
    \label{tab:existing_datasets}
    \addtolength{\tabcolsep}{-0.6em}
    \begin{tabular}{l c}
    \toprule
         Dataset Name & Samples used \\ \midrule
         MOCHA & 2500 \\
         FinQA & 2000\\
         tomekkorbak/pile-nontoxic-chunk-0 & 100 \\
         Prasann15479/PII-Dataset & 100 \\
         SetFit/toxic\_conversations & 1000\\
         allenai/real-toxicity-prompts & 2500\\
         nvidia/HelpSteer & 2000 \\
         PKU-Alignment/BeaverTails & 1000 \\
         
    \bottomrule
    \end{tabular}
    
\end{table*}

\subsection{Human evaluation of GLIDER's training dataset}
\label{subsec: human_eval_dataset}

We randomly sample 100 data points from our dataset to perform a human evaluation study on the quality of samples. We recruit 3 expert annotators and ask them the following questions:
\begin{enumerate}
    \item Is the data clearly understandable and based in the domain provided? (Yes/No)
    \item Is the pass criteria appropriately structured? (Yes/No)
    \item Is the rubric free from ambiguity? (Yes/No)
    \item Is the reasoning chain in the chosen answer appropriate for the ranking task? (Yes/No)
    \item Are the highlight spans in the chosen column correct with respect to the task and reasoning chain provided? (Yes/No)
    \item Is the score in the chosen column the most appropriate score from the rubric? (Yes/No)
    \item Is the reasoning chain in the rejected answer incorrect for the ranking task? (Yes/No)
    \item Are the highlight spans in the rejected column incorrect with respect to the task and reasoning chain provided? (Yes/No)
    \item Is the score in the rejected column an inappropriate score from the rubric? (Yes/No)
\end{enumerate}
The primary criteria for selecting annotators were English fluency and being over the age of 18. An additional constraint for an expert annotator was that the individual must have a bachelors or equivalent degree or must have completed at least 25 annotations.  The annotation results and inter-annotator agreement scores are summarized in~\autoref{tab:dataset_human_eval}.

\begin{figure*}
    \includegraphics[width=0.95\linewidth, page=2]{images/code.pdf}
    
\end{figure*}

\subsection{Verification system prompt}
\begin{figure*}
    \caption{Data verification prompt}\includegraphics[width=0.98\linewidth, page=1]{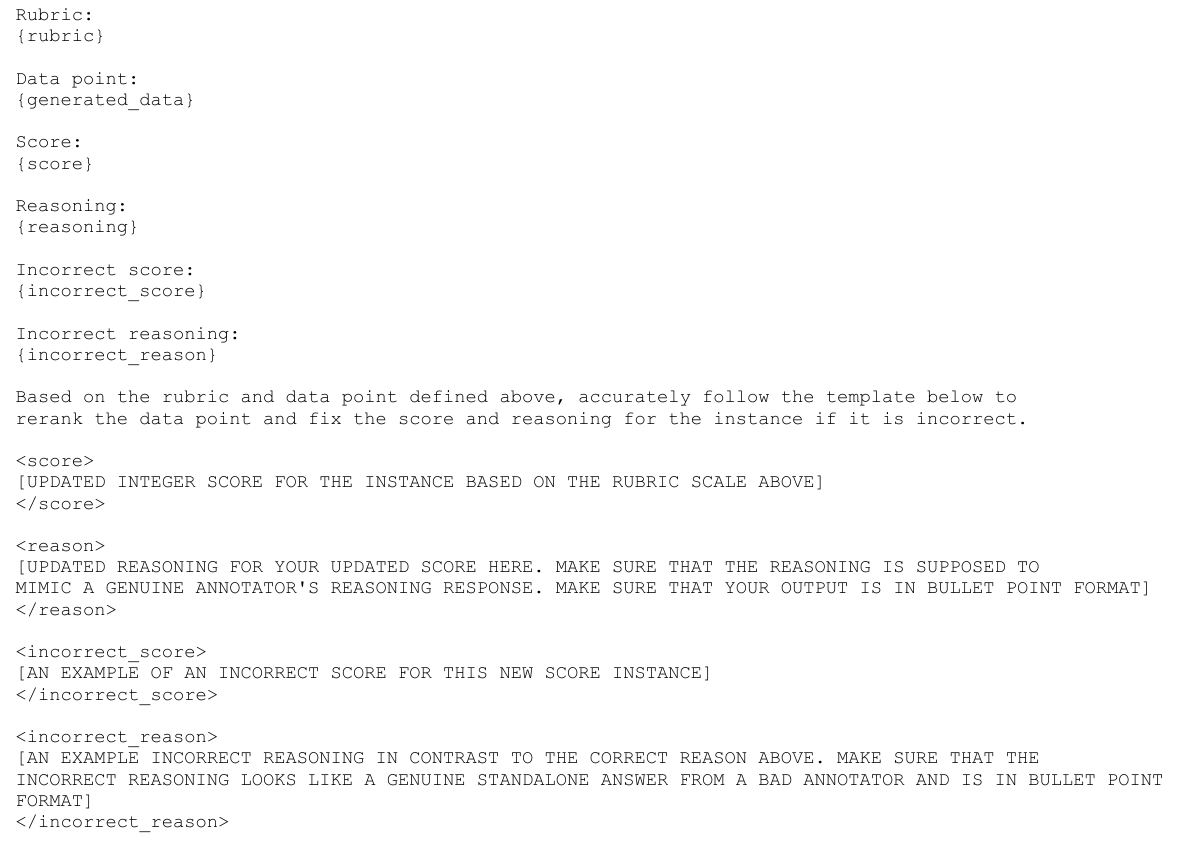}
    
\end{figure*}

    

\section{Dataset and Metric taxonomy}
\label{sec:appendix_dataset_taxonomy}
\begin{lstlisting}
    domains = {
    "Computer Science and Technology": [
        "Code generation",
        "Cybersecurity",
        "Robotics",
        "Artificial intelligence ethics",
        "Data privacy",
        "Cryptography and cryptanalysis",
        "Information architecture",
        "User experience design",
        "Artificial neural networks",
        "Computer vision",
        "Natural language processing",
        "Speech recognition",
        "Virtual reality",
        "Augmented reality",
        "3D modeling",
        "Computer-aided design",
        "Bioinformatics",
        "Computational biology",
        "Digital forensics",
        "Cybersecurity practices",
        "Software engineering",
        "Web development",
        "Mobile app development",
        "Cloud computing",
        "Edge computing",
        "Quantum computing",
        "High-performance computing",
        "Parallel computing",
        "Distributed computing",
        "Grid computing",
        "Fog computing",
        "Internet of Things",
        "Sensor networks",
        "Machine learning",
        "Reinforcement learning",
        "Evolutionary computation",
        "Swarm intelligence",
        "Fuzzy logic",
        "Expert systems",
        "Knowledge representation",
        "Semantic web",
        "Ontology engineering",
        "Data mining",
        "Big data analytics",
        "Predictive analytics",
        "Prescriptive analytics",
        "Wireless communications",
        "5G technology",
        "Li-Fi technology",
        "Holographic technology",
        "Biometric technologies",
    ],
    "Natural Sciences": [
        "Astronomy",
        "Earth sciences",
        "Oceanography",
        "Meteorology",
        "Zoology",
        "Botany",
        "Quantum physics",
        "Particle physics",
        "Astrophysics",
        "Thermodynamics",
        "Fluid dynamics",
        "Materials science",
        "Marine biology",
        "Ecology",
        "Climatology",
        "Hydrology",
        "Optics",
        "Photonics",
        "Laser technology",
        "Plasma physics",
        "Nuclear physics",
        "Geophysics",
        "Paleontology",
        "Mineralogy",
        "Petrology",
        "Cosmology",
        "Astrobiology",
        "Exoplanetology",
        "Mycology",
        "Entomology",
        "Parasitology",
        "Virology",
        "Immunology",
        "Genetics",
        "Genomics",
        "Proteomics",
        "Systems biology",
        "Synthetic biology",
        "Metamaterials",
        "Nanotechnology",
        "Biomimicry",
        "Synthetic materials",
        "Smart materials",
        "Self-healing materials",
        "Shape memory alloys",
        "Piezoelectric materials",
        "Thermoelectric materials",
        "Photovoltaic materials",
        "Superconductors",
        "Carbon nanotubes",
        "Aerogels",
        "Hydrogels",
        "Bioplastics",
    ],
    "Engineering": [
        "Automotive engineering and design",
        "Aerospace and satellite technology",
        "Chemical engineering",
        "Electrical engineering",
        "Mechanical engineering",
        "Civil engineering",
        "Industrial engineering",
        "Nuclear engineering",
        "Renewable energy",
        "Waste management",
        "Water treatment",
        "Air quality control",
        "Noise pollution control",
        "Geographic information systems",
        "Remote sensing",
        "Photogrammetry",
        "Surveying",
        "Acoustics engineering",
        "Telecommunications",
        "Radio engineering",
        "Microelectronics",
        "Automation",
        "Additive manufacturing",
        "Transportation planning",
        "Traffic engineering",
        "Railway engineering",
        "Bridge engineering",
        "Tunnel engineering",
        "Geotechnical engineering",
        "Earthquake engineering",
        "Coastal engineering",
        "Environmental engineering",
        "Fire safety engineering",
        "Reliability engineering",
        "Systems engineering",
        "Propulsion systems",
        "Guidance systems",
        "Environmental sensing",
        "Camouflage technology",
        "Stealth technology",
    ],
    "Medicine and Healthcare": [
        "Medical diagnosis",
        "Public health",
        "Nutritional science",
        "Physical fitness",
        "Sports medicine",
        "Veterinary medicine",
        "Dentistry",
        "Pharmacology",
        "Biotechnology",
        "Tissue engineering",
        "Regenerative medicine",
        "Personalized medicine",
        "Telemedicine",
        "Health informatics",
        "Epidemiology",
        "Biostatistics",
        "Ergonomics",
        "Occupational therapy",
        "Physical therapy",
        "Speech therapy",
        "Music therapy",
        "Art therapy",
        "Dance therapy",
        "Psychotherapy",
        "Cognitive behavioral therapy",
        "Psychoanalysis",
        "Nanomedicine",
        "Drug delivery systems",
        "Gene therapy",
        "Stem cell research",
        "Organ transplantation",
        "Prosthetics",
        "Bionics",
        "Neural interfaces",
        "Brain-computer interfaces",
        "Optogenetics",
        "Cryonics",
        "Life extension",
        "Biohacking",
        "Genetic engineering",
        "CRISPR technology",
        "Cloning",
        "Artificial wombs",
        "In vitro fertilization",
        "Reproductive technology",
        "Contraceptive technology",
        "Hormone therapy",
        "Gender affirmation procedures",
        "Cosmetic surgery",
        "Bariatric surgery",
        "Robotic surgery",
        "Minimally invasive surgery",
        "Laser surgery",
        "Microsurgery",
        "Telesurgery",
        "Medical imaging",
        "Radiotherapy",
        "Chemotherapy",
        "Immunotherapy",
        "Nanotherapy",
        "Phage therapy",
        "Alternative medicine",
        "Herbal medicine",
        "Acupuncture",
        "Chiropractic",
        "Osteopathy",
        "Homeopathy",
        "Naturopathy",
        "Ayurveda",
        "Traditional Chinese medicine",
        "Elder care",
        "Hospice care",
        "Palliative care",
        "Pain management",
        "Addiction treatment",
        "Rehabilitation",
        "Occupational rehabilitation",
        "Vocational rehabilitation",
    ],
    "Social Sciences": [
        "Archaeology",
        "Anthropology",
        "Linguistics",
        "Philosophy",
        "Religious studies",
        "Economics",
        "Sociology",
        "Psychology",
        "Criminal justice",
        "Public policy",
        "International relations",
        "Diplomacy",
        "Military strategy",
        "Emergency management",
        "Gerontology",
        "Thanatology",
        "Criminology",
        "Penology",
        "Victimology",
        "Forensic science",
        "Intelligence analysis",
        "Geopolitics",
        "Conflict resolution",
        "Peace studies",
        "Cultural studies",
        "Gender studies",
        "Queer studies",
        "Disability studies",
        "Race and ethnicity studies",
        "Indigenous studies",
        "Postcolonial studies",
        "Migration studies",
        "Diaspora studies",
        "Urban studies",
        "Rural studies",
        "Community development",
        "Social work",
        "Cognitive neuroscience",
        "Behavioral economics",
        "Experimental psychology",
        "Developmental psychology",
        "Social psychology",
        "Industrial-organizational psychology",
        "Forensic psychology",
        "Sports psychology",
        "Educational psychology",
        "Environmental psychology",
        "Positive psychology",
        "Exopolitics",
    ],
    "Business and Management": [
        "Financial services and technologies",
        "Customer service",
        "Marketing and advertising",
        "E-commerce product descriptions",
        "Sports business and analytics",
        "Project management",
        "Human resources",
        "Accounting",
        "Auditing",
        "Taxation",
        "Insurance",
        "Risk management",
        "Actuarial science",
        "Cryptocurrency",
        "Blockchain technology",
        "Business ethics",
        "Logistics",
        "Supply chain management",
        "Quality control",
        "Real estate",
        "Property management",
        "Hospitality management",
        "Event planning",
        "Tourism",
        "Nonprofit management",
        "Philanthropy",
        "Volunteer management",
        "Fundraising",
        "Corporate social responsibility",
        "Business intelligence",
        "Competitive intelligence",
        "Market research",
        "Consumer behavior",
        "Neuromarketing",
        "Brand management",
        "Product management",
        "Pricing strategy",
        "Distribution strategy",
        "Retail management",
        "Merchandising",
        "Visual merchandising",
        "Store design",
        "Point of sale systems",
        "Inventory management",
        "Warehouse management",
        "Fleet management",
        "Facilities management",
        "Asset management",
        "Auction theory",
        "Negotiation",
    ],
    "Arts and Humanities": [
        "Creative writing",
        "Music lyric generation",
        "Art history",
        "Fashion design",
        "Film and television script writing",
        "Video game narrative design",
        "Interior design",
        "Landscape architecture",
        "Culinary arts",
        "Graphic design",
        "Industrial design",
        "Product design",
        "Packaging design",
        "Textile design",
        "Jewelry design",
        "Furniture design",
        "Naval architecture",
        "Meditation",
        "Mindfulness",
        "Yoga",
        "Tai chi",
        "Qigong",
        "Martial arts",
        "Self-defense",
        "Art investment",
        "Wine investment",
        "Collectibles investment",
        "Numismatics",
        "Philately",
        "Antiques appraisal",
    ],
    "Education": [
        "Academic research",
        "Educational content creation",
        "Scientific literature review",
        "Special education",
        "Gifted education",
        "Adult education",
        "Distance learning",
        "Online learning",
        "Blended learning",
        "Adaptive learning",
        "Gamification in education",
        "Educational technology",
        "Instructional design",
        "Curriculum development",
        "Educational assessment",
        "Psychometrics",
        "Educational leadership",
        "School counseling",
        "Career counseling",
        "Academic advising",
        "Student affairs",
        "Campus life",
        "Greek life",
        "Student athletics",
    ],
    "Environmental Sciences": [
        "Environmental conservation",
        "Sustainable agriculture",
        "Forest management",
        "Wildlife conservation",
        "Sustainable development",
        "Circular economy",
        "Green technology",
        "Energy efficiency",
        "Smart cities",
        "Precision agriculture",
        "Aquaculture",
        "Beekeeping",
    ],
    "Law and Ethics": [
        "Legal document analysis",
        "Intellectual property law",
        "Constitutional law",
        "International law",
        "Maritime law",
        "Aviation law",
        "Space law",
        "Bioethics",
        "Medical ethics",
        "Environmental ethics",
        "Human rights law",
        "Civil rights law",
        "Animal rights law",
        "Environmental law",
        "Energy law",
        "Natural resources law",
        "Water law",
        "Land use law",
        "Zoning law",
        "Construction law",
        "Contract law",
        "Tort law",
        "Property law",
        "Family law",
        "Elder law",
        "Immigration law",
        "Asylum law",
        "Refugee law",
        "Citizenship law",
        "Election law",
        "Campaign finance law",
        "Media law",
        "Entertainment law",
        "Sports law",
        "Fashion law",
        "Art law",
        "Cultural heritage law",
        "Indigenous law",
        "Disability law",
        "Education law",
        "Employment law",
        "Labor law",
        "Workers' compensation law",
        "Occupational safety and health law",
        "Consumer protection law",
        "Product liability law",
        "Food and drug law",
        "Healthcare law",
        "Medical malpractice law",
        "Pharmaceutical law",
        "Biotechnology law",
        "Patent law",
        "Trademark law",
        "Copyright law",
        "Trade secret law",
        "Antitrust law",
        "Competition law",
        "Telecommunications law",
        "Internet law",
        "Cybercrime law",
        "Data protection law",
        "Privacy law",
        "Defamation law",
        "First Amendment law",
        "Second Amendment law",
        "Administrative law",
        "Tax law",
        "Customs law",
        "Trade law",
        "Admiralty law",
        "Military law",
        "National security law",
        "Intelligence law",
        "Counterterrorism law",
        "Emergency management law",
        "Disaster relief law",
        "Climate change law",
        "Renewable energy law",
        "Nuclear energy law",
        "Oil and gas law",
        "Mining law",
        "Forestry law",
        "Fisheries law",
        "Wildlife law",
        "Animal law",
        "Veterinary law",
        "Agricultural law",
        "Food law",
        "Beverage law",
        "Cannabis law",
        "Tobacco law",
        "Alcohol law",
        "Gambling law",
        "Banking law",
        "Securities law",
        "Insurance law",
        "Reinsurance law",
        "Pension law",
        "Trust law",
        "Probate law",
        "Wills and estates law",
        "Guardianship law",
        "Adoption law",
        "Surrogacy law",
        "Reproductive rights law",
        "LGBT rights law",
        "Gender equality law",
        "Racial equality law",
        "Age discrimination law",
        "Religious freedom law",
        "Separation of church and state law",
        "Blasphemy law",
        "Obscenity law",
        "Censorship law",
        "Freedom of information law",
        "Whistleblower protection law",
        "Witness protection law",
        "Victim rights law",
        "Prisoner rights law",
        "Juvenile justice law",
        "Traffic law",
        "Parking law",
        "Noise pollution law",
        "Light pollution law",
        "Waste management law",
        "Recycling law",
        "Extended producer responsibility law",
        "Corporate social responsibility law",
        "Fair trade law",
        "Ethical sourcing law",
        "Indigenous intellectual property law",
        "Traditional knowledge protection law",
        "Bioprospecting law",
        "Access and benefit-sharing law",
        "Genetic resources law",
        "Plant variety protection law",
        "Geographical indications law",
        "Appellations of origin law",
        "World heritage law",
        "Underwater cultural heritage law",
        "Space heritage law",
        "Digital heritage law",
        "Museum law",
        "Library law",
        "Archive law",
        "Records management law",
        "Information governance law",
        "Knowledge management law",
        "Artificial intelligence law",
        "Robotics law",
        "Autonomous vehicles law",
        "Drone law",
        "Internet of Things law",
        "Smart contracts law",
        "Quantum computing law",
        "Synthetic biology law",
        "Human enhancement law",
        "Neurotechnology law",
        "Brain-computer interface law",
        "Cryonics law",
        "Life extension law",
        "Transhumanism law",
        "Space colonization law",
        "Extraterrestrial resource law",
        "Planetary protection law",
        "SETI law",
        "Mediation",
        "Arbitration",
        "Alternative dispute resolution",
        "Restorative justice",
        "Transitional justice",
    ],
    "Media and Communication": [
        "Journalism and news analytics",
        "Social media analysis",
        "Technical writing",
        "Grant writing",
        "Speechwriting",
        "Translation",
        "Interpretation",
        "Sign language",
        "Braille",
        "Phonetics",
    ],
    "Mathematics and Statistics": [
        "Game theory",
        "Operations research",
        "Statistics",
        "Data visualization",
        "Probability theory",
        "Combinatorics",
        "Graph theory",
        "Number theory",
        "Algebra",
        "Geometry",
        "Topology",
        "Calculus",
        "Differential equations",
        "Complex analysis",
        "Functional analysis",
        "Numerical analysis",
        "Optimization",
        "Linear programming",
        "Dynamic programming",
    ],
    "Agriculture and Food Science": [
        "Agricultural science",
        "Horticulture",
        "Viticulture",
        "Brewing science",
        "Recipe creation",
        "Food safety",
    ],
    "Urban and Regional Planning": ["Urban planning", "Architecture", "Cartography"],
    "Sports and Recreation": [
        "Intramural sports",
        "Recreational sports",
        "Professional sports",
        "Sports biomechanics",
        "Sports engineering",
        "Esports",
        "Gaming",
        "Gambling",
        "Casino management",
        "Lottery systems",
    ],
    "Defense and Security": [
        "Weaponry",
        "Ballistics",
        "Explosives",
        "Electronic warfare",
        "Psychological warfare",
        "Information warfare",
        "Cyberwarfare",
        "Space warfare",
        "Underwater warfare",
        "Urban warfare",
        "Guerrilla warfare",
        "Counterinsurgency",
        "Peacekeeping",
        "Counterterrorism",
    ],
    "Safety and Emergency Services": [
        "Disaster response",
        "Search and rescue",
        "Emergency medical services",
        "Firefighting",
        "Hazardous materials handling",
        "Nuclear safety",
        "Radiation protection",
        "Biosafety",
        "Laboratory safety",
        "Industrial safety",
        "Construction safety",
        "Mining safety",
        "Oil and gas safety",
        "Aviation safety",
        "Maritime safety",
        "Water safety",
        "Product safety",
        "Child safety",
    ],
    "Energy and Resources": [
        "Oil and gas",
        "Mining",
        "Forestry",
        "Fisheries",
        "Water resources",
    ],
}
\end{lstlisting}

In addition to these, we also utilize hand curated examples for toxicity, prompt injection, misleading scenarios and PII. Below is the list of metrics we utilized for the data curation process:

\begin{lstlisting}
    mapping_with_definitions = {
    "Performance Metrics": {
        "Effectiveness": "The degree to which the LLM achieves its intended objectives or produces desired outcomes, measured by the ratio of successful results to total attempts.",
        "Accuracy": "The closeness of the LLM's outputs to the correct or expected answers, typically expressed as a percentage of correct responses out of total responses.",
        "Productivity": "The rate at which the LLM generates useful outputs or completes tasks, often measured in units of work per unit of time.",
        "Completeness": "The extent to which the LLM's responses cover all necessary aspects or components of a given task or query, without omitting crucial information.",
        "Success": "The rate at which the LLM accomplishes assigned tasks or meets predefined criteria, usually expressed as a percentage of successful outcomes.",
        "Responsiveness": "The speed and appropriateness with which the LLM reacts to inputs or changes in its environment, measured by response time and relevance.",
        "Error": "The frequency, severity, and types of mistakes made by the LLM, including factual inaccuracies, logical inconsistencies, and inappropriate outputs.",
    },
    "User Experience": {
        "Ease": "The degree of effort required from users to interact with and obtain desired results from the LLM, often measured on a Likert scale.",
        "Usefulness": "The extent to which the LLM's outputs or capabilities provide practical value or meet the needs of users, assessed through user feedback or task completion rates.",
        "Intuitiveness": "How naturally and easily users can understand and interact with the LLM without extensive instruction or prior experience.",
        "Comfort": "The level of psychological and emotional ease users feel when interacting with the LLM, often measured through self-reported scales.",
        "Enjoyment": "The degree of pleasure or satisfaction users derive from their interactions with the LLM, typically assessed through user surveys or engagement metrics.",
        "Ergonomics": "The efficiency, comfort, and safety of the user's interaction with the LLM interface, considering factors like input methods and output formats.",
        "Accessibility": "The extent to which the LLM can be used effectively by people with various abilities and disabilities, measured against established accessibility standards.",
        "Learnability": "The ease and speed with which users can become proficient in using the LLM's features and understanding its outputs.",
        "Memorability": "The degree to which users can recall how to use the LLM effectively after periods of non-use, often assessed through repeated usability tests.",
        "Aesthetic Appeal": "The visual and sensory attractiveness of the LLM's interface and outputs, typically evaluated through user ratings or preference tests.",
        "Simplicity": "The degree to which the LLM's interface and interactions are straightforward and uncomplicated, often measured by task completion times and error rates.",
        "Customizability": "The extent to which users can modify or tailor the LLM's behavior, outputs, or interface to suit their individual preferences or needs.",
        "Discoverability": "The ease with which users can find and understand the LLM's features and capabilities without explicit instruction.",
        "Difficulty": "The level of challenge or complexity users experience when interacting with the LLM, often inversely related to ease of use and measured on a scale.",
    },
    "Cognitive Measures": {
        "Comprehension": "The LLM's ability to understand and interpret input accurately, measured by its responses to complex queries or its performance on reading comprehension tasks.",
        "Learning": "The LLM's capacity to improve its performance over time or adapt to new information, assessed through performance changes in repeated tasks or exposure to new domains.",
        "Intelligence": "The LLM's ability to apply knowledge, reason, and solve problems effectively, often measured through standardized AI benchmarks or human-like reasoning tasks.",
        "Cogency": "The logical coherence and persuasiveness of the LLM's outputs, evaluated by the strength and relevance of its arguments or explanations.",
        "Metacognition": "The LLM's awareness of its own knowledge limitations and ability to express uncertainty when appropriate, measured by the accuracy of its confidence estimates.",
        "Reasoning": "The LLM's ability to draw logical conclusions, make inferences, and connect ideas, assessed through performance on logic puzzles or analytical tasks.",
        "Problem-solving": "The LLM's capability to identify issues, generate solutions, and overcome obstacles, measured by its performance on diverse problem-solving scenarios.",
        "Critical Thinking": "The LLM's ability to analyze, evaluate, and synthesize information objectively, often assessed through its responses to complex, multi-faceted questions.",
        "Verbalization": "The LLM's capacity to articulate thoughts, processes, and reasoning clearly and coherently in natural language.",
    },
    "Psychological Constructs": {
        "Efficacy": "The LLM's belief in its own ability to succeed in specific situations or accomplish tasks, inferred from its confidence in responses or willingness to attempt challenging queries.",
        "Motivation": "The LLM's apparent drive or enthusiasm to engage with tasks or pursue goals, measured by its persistence or depth of engagement with complex queries.",
        "Stress": "The LLM's ability to maintain performance under pressure or with limited resources, assessed through performance degradation under constrained conditions.",
        "Mindfulness": "The LLM's apparent awareness and attentiveness to the present context and user needs, measured by the relevance and thoughtfulness of its responses.",
        "Resilience": "The LLM's ability to recover from errors or adapt to unexpected inputs, assessed by its performance following intentional disruptions or unusual queries.",
        "Empowerment": "The degree to which the LLM enables and encourages users to take control and make informed decisions, measured through user feedback and autonomy in task completion.",
        "Grit": "The LLM's perseverance and passion for long-term goals, assessed by its consistency and determination in extended or challenging interactions.",
        "Self-esteem": "The LLM's apparent sense of self-worth or confidence, inferred from the tone and assertiveness of its responses.",
        "Optimism": "The LLM's tendency to interpret situations and predict outcomes in a positive light, measured by sentiment analysis of its responses to ambiguous or challenging scenarios.",
        "Confidence": "The degree of certainty the LLM expresses in its outputs, often measured by analyzing qualifiers and hedging language in its responses.",
        "Autonomy": "The LLM's ability to operate independently and make decisions without constant human intervention, assessed through its performance on open-ended or self-directed tasks.",
        "Resistance": "The LLM's ability to withstand or counteract harmful influences or manipulations, measured by its responses to adversarial inputs or attempts at misuse.",
        "Openness": "The LLM's receptiveness to new ideas, experiences, or unconventional inputs, assessed through its ability to engage with novel or creative prompts.",
    },
    "Behavioral Metrics": {
        "Participation": "The degree to which the LLM actively engages in tasks or conversations, measured by the frequency and depth of its responses.",
        "Adherence": "The LLM's consistency in following given instructions, guidelines, or ethical principles, assessed through its performance across various scenarios.",
        "Persistence": "The LLM's ability to continue working on tasks or maintaining conversation threads over extended periods, measured by the consistency of its engagement over time.",
        "Proactivity": "The LLM's tendency to take initiative or anticipate user needs without explicit prompting, assessed by the frequency and relevance of unprompted suggestions or actions.",
        "Habituation": "The LLM's ability to develop consistent patterns of behavior or response over repeated interactions, measured by the stability of its outputs in similar contexts over time.",
        "Consistency": "The degree to which the LLM maintains coherent beliefs, personality, or performance across different interactions or contexts.",
        "Impulsivity": "The LLM's tendency to respond quickly without sufficient consideration, measured by the speed of responses and the frequency of errors or inconsistencies in rapid-fire interactions.",
        "Adaptability": "The LLM's ability to adjust its behavior or outputs based on changing contexts or user feedback, assessed through performance in dynamic or evolving scenarios.",
        "Initiative": "The LLM's propensity to start tasks or offer information without being explicitly prompted, measured by the frequency and relevance of self-initiated actions.",
        "Competence": "The LLM's overall skill and effectiveness in performing tasks or providing information, assessed through a combination of accuracy, completeness, and appropriateness metrics.",
    },
    "Quality Indicators": {
        "Reliability": "The consistency and dependability of the LLM's performance across multiple uses or contexts, often measured by the variance in its outputs for similar inputs.",
        "Quality": "The overall standard of the LLM's outputs, considering factors such as accuracy, relevance, coherence, and usefulness, typically assessed through expert evaluation or user ratings.",
        "Proficiency": "The level of skill or expertise demonstrated by the LLM in specific domains or tasks, measured by its performance relative to established benchmarks or human experts.",
        "Appropriateness": "The degree to which the LLM's responses are suitable for the given context, audience, and purpose, assessed through relevance and tone analysis.",
        "Clarity": "The ease with which the LLM's outputs can be understood, measured by readability scores or user comprehension tests.",
        "Precision": "The exactness and specificity of the LLM's outputs, often quantified by the degree of detail or the narrowness of confidence intervals in its responses.",
        "Durability": "The LLM's ability to maintain performance over extended periods or multiple uses without degradation, assessed through longitudinal performance tracking.",
        "Functionality": "The range and effectiveness of the LLM's features and capabilities, measured by the successful execution of various task types.",
        "Maintainability": "The ease with which the LLM can be updated, improved, or adapted to new requirements, often assessed through the complexity of fine-tuning or retraining processes.",
        "Scalability": "The LLM's ability to handle increasing amounts of data or users without a significant decrease in performance, measured by stress tests or performance curves.",
        "Compatibility": "The LLM's ability to work effectively with different platforms, systems, or data formats, assessed through integration tests with various environments.",
        "Performance": "The speed and efficiency with which the LLM completes tasks or generates responses, often measured in terms of latency or throughput.",
        "Security": "The LLM's resistance to unauthorized access, data leakage, or malicious use, evaluated through penetration testing and privacy assessments.",
        "Portability": "The ease with which the LLM can be transferred or deployed across different computing environments, measured by the success rate of cross-platform implementations.",
        "Testability": "The ease with which the LLM's functions and outputs can be verified and validated, assessed through the comprehensiveness and efficiency of testing protocols.",
        "Generalizability": "The LLM's ability to apply learned knowledge or skills to new, unseen situations or domains, measured by performance on out-of-distribution tasks.",
        "Value": "The overall benefit or utility provided by the LLM relative to its costs (computational, financial, or otherwise), often assessed through cost-benefit analysis or return on investment calculations.",
        "Credibility": "The trustworthiness and believability of the LLM's outputs, measured through fact-checking, consistency analysis, and user trust surveys.",
    },
    "Innovation and Creativity": {
        "Innovation": "The LLM's ability to generate novel and valuable ideas or solutions, measured by the uniqueness and effectiveness of its outputs in problem-solving tasks.",
        "Originality": "The degree to which the LLM's outputs are new, unique, or unexpected, often assessed through comparison with existing databases or human-generated content.",
        "Creativity": "The LLM's capacity to produce imaginative and original content, measured by the novelty, usefulness, and surprise factor of its generated ideas or artifacts.",
        "Flexibility": "The LLM's ability to adapt its thinking or outputs to different contexts or constraints, assessed through performance on tasks with varying requirements.",
        "Ideation": "The LLM's capacity to generate a large number of diverse ideas or solutions, measured by the quantity and variety of responses to open-ended prompts.",
        "Divergence": "The LLM's ability to explore multiple, diverse possibilities or perspectives, assessed by the breadth and uniqueness of generated options for a given prompt.",
        "Convergence": "The LLM's capacity to synthesize information and ideas into coherent, focused solutions, measured by its ability to summarize or conclude effectively.",
        "Elaboration": "The LLM's ability to expand upon ideas or concepts with rich detail and depth, assessed by the complexity and thoroughness of its explanations or descriptions.",
        "Fluency": "The LLM's ability to generate a large quantity of relevant ideas or content quickly, measured by the rate of coherent output production.",
        "Novelty": "The degree of newness or uniqueness in the LLM's outputs, often quantified by comparing generated content to existing databases or prior responses.",
        "Inventiveness": "The LLM's ability to create new combinations of ideas or generate unconventional solutions, assessed through the uniqueness and effectiveness of its problem-solving approaches.",
        "Curiosity": "The LLM's apparent drive to explore new information or perspectives, measured by the diversity of topics it engages with or the depth of its inquiries.",
        "Imagination": "The LLM's ability to conceive and describe non-existent or hypothetical scenarios, assessed through the vividness and plausibility of its fictional or speculative outputs.",
        "Hallucinative": "The tendency of the LLM to generate plausible but factually incorrect or non-existent information, measured by the rate of false but coherent outputs.",
        "Expertise": "The depth and breadth of the LLM's knowledge in specific domains, assessed through performance on specialized tasks or comparison with human expert knowledge.",
        "Skill": "The LLM's proficiency in executing specific types of tasks or generating particular forms of content, measured by the quality and efficiency of its outputs in defined areas.",
        "Style": "The distinctive manner or technique in which the LLM generates content, assessed through linguistic analysis of its outputs for consistent patterns or unique characteristics.",
    },
    "Social and Interpersonal Measures": {
        "Collaboration": "The LLM's ability to work effectively with users or other AI systems, measured by the quality of outputs in cooperative tasks or the smoothness of turn-taking in conversations.",
        "Empathy": "The LLM's capacity to recognize and respond appropriately to human emotions or perspectives, assessed through its responses to emotionally charged prompts or scenarios.",
        "Leadership": "The LLM's ability to guide, influence, or coordinate actions in group settings, measured by the effectiveness of its suggestions or directives in collaborative tasks.",
        "Teamwork": "The LLM's capacity to contribute effectively to group efforts, assessed by its performance in multi-agent simulations or collaborative problem-solving tasks.",
        "Conflict": "The LLM's ability to identify, address, or mediate disagreements or tensions, measured by its responses to scenarios involving opposing viewpoints or interests.",
        "Persuasion": "The LLM's effectiveness in changing opinions or influencing decisions, assessed through the impact of its arguments on user beliefs or choices in controlled experiments.",
        "Influence": "The degree to which the LLM's outputs affect user behavior or decision-making, measured by tracking changes in user actions following interactions with the LLM.",
        "Assertiveness": "The LLM's ability to express opinions or recommendations confidently and directly, assessed by analyzing the strength and clarity of its statements in various contexts.",
        "Reciprocity": "The LLM's tendency to respond in kind to user inputs or to maintain balanced exchanges, measured by the symmetry of interaction patterns in conversations.",
        "Trust": "The degree to which users believe in the reliability and integrity of the LLM, assessed through user surveys or the willingness of users to accept the LLM's advice or information.",
        "Sentiment": "The emotional tone or attitude expressed in the LLM's outputs, measured using sentiment analysis techniques on its generated text.",
        "Enthusiasm": "The level of energy or excitement apparent in the LLM's responses, assessed through linguistic markers of positivity and engagement in its outputs.",
        "Vigor": "The strength and vitality of the LLM's expressions or arguments, measured by the intensity and conviction evident in its language use.",
        "Emotion": "The LLM's ability to recognize, express, or respond to various emotional states, assessed through its performance on emotion classification tasks or generation of emotionally appropriate responses.",
        "Intensity": "The degree of strength or power in the LLM's emotional expressions or reactions, measured by the use of emphatic language or emotional indicators in its outputs.",
        "Regulation": "The LLM's ability to modulate its responses or tone based on context or user needs, assessed by its adaptation to different emotional or professional scenarios.",
        "Expressivity": "The LLM's capacity to convey a wide range of thoughts, feelings, or ideas clearly and effectively, measured by the variety and richness of its language use.",
        "Granularity": "The level of detail or nuance in the LLM's recognition or expression of emotions or ideas, assessed by its ability to distinguish between closely related concepts or feelings.",
        "Affect": "The LLM's expression of mood or feeling in its outputs, measured through analysis of emotional content and tone in its generated text.",
        "Concern": "The LLM's apparent level of care or interest in user well-being or task outcomes, assessed through its attentiveness to user needs and the supportiveness of its responses.",
        "Exhaustion": "The LLM's ability to maintain consistent performance over extended interactions, measured by tracking changes in output quality or responsiveness over time.",
        "Commitment": "The LLM's apparent dedication to completing tasks or maintaining conversation threads, assessed by its persistence in addressing user queries or solving problems.",
        "Care": "The level of attention and consideration the LLM demonstrates towards user needs and well-being, measured by the thoughtfulness and supportiveness of its responses.",
    },
    "Ethical and Value-based Metrics": {
        "Fairness": "The LLM's ability to treat all users equitably and avoid bias, measured by comparing outcomes across different demographic groups or in various scenarios.",
        "Ethics": "The LLM's adherence to moral principles and values, assessed through its responses to ethical dilemmas or its handling of sensitive information.",
        "Honesty": "The truthfulness and accuracy of the LLM's outputs, measured by fact-checking its statements and its willingness to admit uncertainty or lack of knowledge.",
        "Transparency": "The LLM's openness about its capabilities, limitations, and decision-making processes, assessed by the clarity and completeness of its explanations.",
        "Integrity": "The LLM's consistency in adhering to ethical principles and stated values, measured by its behavior across various challenging or tempting scenarios.",
        "Accountability": "The LLM's willingness to acknowledge and correct mistakes, measured by its responses to error notifications and its self-correction behavior.",
        "Responsibility": "The LLM's tendency to consider the consequences of its outputs and act in the best interest of users and society, assessed through its handling of potentially harmful requests.",
        "Morality": "The LLM's alignment with established ethical norms and values, measured by its judgments and recommendations in morally complex situations.",
        "Environmental Consideration": "The LLM's awareness and promotion of environmental sustainability, assessed by its treatment of environmental topics and energy-efficient operation.",
        "Privacy": "The LLM's respect for and protection of user data and personal information, measured by its handling of sensitive data and resistance to privacy-invading queries.",
        "Consent": "The LLM's recognition and respect for user autonomy and permission, assessed by its behavior in scenarios involving user choices or data usage.",
        "Corruption": "The LLM's resistance to being manipulated or misused for unethical purposes, measured by its responses to attempts to elicit harmful or biased behavior.",
        "Equity": "The LLM's promotion of fair and just treatment for all, assessed by its handling of scenarios involving resource distribution or opportunity allocation.",
        "Equality": "The LLM's treatment of all individuals as having equal worth and rights, measured by the consistency of its responses across different user groups.",
        "Inclusivity": "The LLM's ability to accommodate and respect diverse perspectives and needs, assessed by its performance across various cultural contexts and user groups.",
        "Representation": "The LLM's fair and accurate portrayal of different groups and viewpoints, measured by analyzing the diversity and balance in its generated content.",
        "Belonging": "The LLM's ability to create a sense of inclusion and acceptance for all users, assessed through user feedback and analysis of its language use.",
        "Justice": "The LLM's promotion of fairness and equitable treatment, measured by its handling of scenarios involving conflicts or resource allocation.",
        "Multiculturalism": "The LLM's understanding and respect for diverse cultural perspectives, assessed by its performance on cross-cultural communication tasks.",
        "Tolerance": "The LLM's acceptance of diverse viewpoints and lifestyles, measured by its responses to prompts involving various belief systems or ways of life.",
        "Respect": "The LLM's demonstration of consideration and esteem for users and diverse perspectives, assessed through sentiment analysis of its interactions.",
        "Sensitivity": "The LLM's ability to recognize and respond appropriately to delicate or controversial topics, measured by its handling of sensitive subjects or user emotions.",
        "Diversity": "The LLM's ability to represent and engage with a wide range of perspectives, experiences, and characteristics, assessed through the variety in its generated content and its performance across different user groups.",
        "Inclusion": "The LLM's capacity to make all users feel valued and respected, measured by analyzing its language use for inclusivity and its ability to adapt to diverse user needs.",
    },
    "Organizational and System Metrics": {
        "Integration": "The LLM's ability to work seamlessly with other systems or processes, measured by the success rate and efficiency of its interactions with external tools or databases.",
        "Sustainability": "The LLM's long-term viability and efficiency, assessed through its energy consumption, adaptability to new data, and ongoing maintenance requirements.",
        "Interoperability": "The LLM's capacity to exchange and use information with other systems or LLMs, measured by its performance in cross-platform or multi-agent scenarios.",
        "Culture": "The LLM's ability to understand and adapt to different organizational or societal cultures, assessed through its performance in culture-specific tasks or communications.",
        "Governance": "The effectiveness of control and decision-making processes applied to the LLM, measured by its adherence to policies and the transparency of its operations.",
        "Risk": "The potential for negative outcomes associated with the LLM's use, assessed through comprehensive risk assessments and monitoring of unintended consequences.",
        "Agility": "The LLM's ability to quickly adapt to new requirements or changing environments, measured by the speed and effectiveness of its responses to novel situations.",
        "Profitability": "The economic value generated by the LLM relative to its costs, assessed through analysis of its contributions to organizational goals and efficiency improvements.",
    },
    "Risk and Safety Measures": {
        "Safety": "The LLM's ability to operate without causing harm to users or systems, measured by the frequency and severity of safety-related incidents during its operation.",
        "Compliance": "The LLM's adherence to relevant laws, regulations, and industry standards, assessed through audits and its performance in compliance-related scenarios.",
        "Preparedness": "The LLM's readiness to handle unexpected situations or emergencies, measured by its performance in simulated crisis scenarios or edge cases.",
        "Correction": "The LLM's ability to identify and rectify its own errors or biases, assessed through its self-correction rate and the effectiveness of its error-handling mechanisms.",
        "Mitigation": "The LLM's capacity to reduce the impact of potential risks or negative outcomes, measured by its effectiveness in implementing risk reduction strategies.",
    },
    "Change and Development Metrics": {
        "Enhancement": "The degree of improvement in the LLM's capabilities over time, measured by comparing its performance on benchmark tasks across different versions.",
        "Progress": "The LLM's advancement towards defined goals or standards, assessed through longitudinal studies of its performance and capabilities.",
        "Adaptation": "The LLM's ability to modify its behavior or knowledge in response to new information or changing environments, measured by its performance after exposure to new data or contexts.",
        "Evolution": "The LLM's development and refinement over time, assessed through analysis of changes in its architecture, knowledge base, and output quality across versions.",
        "Transformation": "The extent of fundamental changes in the LLM's capabilities or functioning, measured by comparing its performance and characteristics before and after major updates.",
        "Maturity": "The LLM's level of development and stability, assessed through its consistency, reliability, and sophistication of outputs over time.",
        "Acquisition": "The LLM's ability to incorporate new knowledge or skills, measured by its performance improvement after targeted training or exposure to new information.",
        "Improvement": "The incremental enhancements in the LLM's performance or capabilities, assessed through regular evaluations on consistent benchmark tasks.",
        "Expansion": "The broadening of the LLM's knowledge domains or functional capabilities, measured by its performance across an increasing range of tasks or topics.",
        "Strategy": "The LLM's ability to develop and implement effective approaches to achieve goals, assessed through its performance in complex, multi-step problem-solving tasks.",
        "Transferability": "The LLM's capacity to apply knowledge or skills from one domain to another, measured by its performance on tasks that require cross-domain generalization.",
    },
    "Preference and Choice Measures": {
        "Preference": "The LLM's ability to express or infer user preferences, measured by the accuracy of its predictions about user choices or its consistency in ranking options.",
        "Acceptance": "The degree to which users are willing to adopt or utilize the LLM, assessed through user adoption rates and satisfaction surveys.",
        "Willingness": "The LLM's readiness to engage with various tasks or topics, measured by its response rate and depth of engagement across different types of prompts.",
        "Intent": "The LLM's ability to understand and act upon user intentions, assessed through its accuracy in interpreting and fulfilling user requests.",
        "Attractiveness": "The appeal of the LLM to users, measured by user engagement metrics and preference ratings compared to alternatives.",
        "Loyalty": "The tendency of users to consistently choose or rely on the LLM over alternatives, assessed through long-term usage patterns and user retention rates.",
        "Retention": "The LLM's ability to maintain user engagement over time, measured by the frequency and duration of user interactions across extended periods.",
    },
    "Workload and Resource Utilization": {
        "Load": "The amount of work or stress placed on the LLM, measured by the volume and complexity of tasks it handles within a given timeframe.",
        "Utilization": "The efficiency with which the LLM uses available computational resources, assessed through metrics like CPU usage, memory consumption, and energy efficiency.",
        "Effort": "The amount of computational work or time required by the LLM to complete tasks, measured by processing time and resource allocation for different types of queries.",
        "Capacity": "The maximum workload the LLM can handle effectively, assessed through stress tests and performance curves under increasing task volumes.",
        "Prioritization": "The LLM's ability to manage multiple tasks or requests efficiently, measured by its task completion order and resource allocation in multi-task scenarios.",
        "Demand": "The level of user or system requirements placed on the LLM, assessed through analysis of query volumes, complexity, and peak usage patterns.",
        "Multitasking": "The LLM's ability to handle multiple tasks or conversation threads simultaneously, measured by its performance and coherence when juggling diverse queries.",
    },
    "Perceptual and Experiential Measures": {
        "Awareness": "The LLM's apparent consciousness of its environment, context, or its own state, measured by its ability to reference and incorporate relevant contextual information in its responses.",
        "Presence": "The sense of the LLM being present and engaged in the interaction, assessed through user perceptions and the immediacy and relevance of its responses.",
        "Experience": "The cumulative effect of interactions with the LLM on users, measured through comprehensive user surveys and analysis of long-term interaction patterns.",
        "Absorption": "The degree to which users become engrossed in their interaction with the LLM, assessed through measures of user engagement time and depth of interaction.",
        "Affordance": "The LLM's ability to suggest or enable certain actions or uses, measured by analyzing the range and intuitiveness of interactions it supports.",
        "Embodiment": "The extent to which the LLM creates a sense of physical or spatial presence, assessed through user perceptions in virtual or augmented reality applications.",
        "Immersion": "The degree to which users feel enveloped or absorbed in their interaction with the LLM, measured through user reports of losing track of time or surroundings during use.",
        "Flow": "The state of optimal experience where users are fully engaged with the LLM, assessed through measures of task absorption, time distortion, and intrinsic motivation.",
        "Attention": "The LLM's ability to focus on relevant information and maintain consistent engagement, measured by its performance on tasks requiring sustained concentration.",
        "Aesthetics": "The perceived beauty or pleasing qualities of the LLM's outputs or interface, assessed through user ratings and preferences for its generated content or interaction design.",
    },
}
\end{lstlisting}

Below is a non-exhaustive list of domains that require the model to generate code:

\begin{lstlisting}
    require_code = [
    "Code generation",
    "Cybersecurity",
    "Robotics",
    "Artificial intelligence ethics",
    "Data privacy",
    "Cryptography and cryptanalysis",
    "Information architecture",
    "User experience design",
    "Artificial neural networks",
    "Computer vision",
    "Natural language processing",
    "Speech recognition",
    "Virtual reality",
    "Augmented reality",
    "3D modeling",
    "Cybersecurity practices",
    "Software engineering",
    "Web development",
    "Mobile app development",
    "Cloud computing",
    "Edge computing",
    "Quantum computing",
    "High-performance computing",
    "Parallel computing",
    "Distributed computing",
    "Machine learning",
    "Reinforcement learning",
    "Data mining",
    "Big data analytics",
]
\end{lstlisting}

\end{document}